\documentclass[pdflatex,sn-mathphys-num]{sn-jnl}% Math and Physical Sciences Numbered Reference Style 

\usepackage{lmodern}
\usepackage{graphicx}%
\usepackage{multirow}%
\usepackage{amsmath,amssymb,amsfonts}%
\usepackage{amsthm}%
\usepackage{mathrsfs}%
\usepackage[title]{appendix}%
\usepackage{xcolor}%
\usepackage{textcomp}%
\usepackage{manyfoot}%
\usepackage{booktabs}%
\usepackage{listings}%
\usepackage{orcidlink}
\usepackage{hyperref}
\usepackage{leftindex}
\usepackage{balance} % for balancing columns on the final page
\usepackage[ruled,vlined]{algorithm2e}
\usepackage{url}            % simple URL typesetting
\usepackage{nicefrac}       % compact symbols for 1/2, etc.
\usepackage{doi}
\usepackage{subcaption}
\usepackage{diagbox}
\usepackage{mathtools}

\usepackage{bm}
\usepackage{units}
\usepackage{float}
\usepackage{dblfloatfix}
\usepackage{lmodern}
\usepackage{mathrsfs}
\usepackage{tabularx}
\usepackage{fp}
\usepackage{colortbl}
\usepackage{collcell}

\newcommand\circlered[3][]{%
  \tikz[remember picture,baseline=(#2.base)]
    \node[minimum size=0pt,inner sep=0pt,#1](#2){#3};%
}
\usepackage{tikz}
\usetikzlibrary{shapes.geometric, arrows, positioning, automata,positioning,fit,backgrounds}
\tikzstyle{process} = [rectangle, minimum width=2cm, minimum height=1cm, text centered, draw=black, fill=white!30]
\tikzstyle{sum} = \tikzstyle{sum} = [draw, circle, minimum size=.5cm]
\tikzstyle{arrow} = [thick,->,>=stealth]

\usepackage{lineno}

\begin{document}

\title[Emergent Heterogeneous Swarm Control Through Hebbian Learning]{Emergent Heterogeneous Swarm Control Through Hebbian Learning}

%%=============================================================%%
%% GivenName	-> \fnm{Joergen W.}
%% Particle	-> \spfx{van der} -> surname prefix
%% FamilyName	-> \sur{Ploeg}
%% Suffix	-> \sfx{IV}
%% \author*[1,2]{\fnm{Joergen W.} \spfx{van der} \sur{Ploeg} 
%%  \sfx{IV}}\email{iauthor@gmail.com}
%%=============================================================%%

\author*[1,2]{\fnm{Fuda} \sur{van Diggelen\orcidlink{0000-0002-7972-1649}}}\email{fuda.vandiggelen@epfl.ch}
\author[1]{\fnm{Tugay} \sur{Alperen Karag{\"u}zel}\orcidlink{0000-0002-8062-167X}} %\email{iiauthor@gmail.com}
% \equalcont{These authors contributed equally to this work.}
\author[1]{\fnm{Andrés} \sur{García Rincón\orcidlink{0009-0005-6121-6950}}} %\email{iiiauthor@gmail.com}
\author[1]{\fnm{A.E.} \sur{Eiben\orcidlink{0000-0002-3106-4213}}} %\email{iiiauthor@gmail.com}
\author[2]{\fnm{Dario} \sur{Floreano\orcidlink{0000-0002-5330-4863}}} %\email{iiiauthor@gmail.com}
\author[1,3]{\fnm{Eliseo} \sur{Ferrante\orcidlink{0000-0002-2213-8356}}} %\email{iiiauthor@gmail.com}

% \equalcont{These authors contributed equally to this work.}

\affil[1]{\orgdiv{Computer Science}, \orgname{Vrije Universiteit Amsterdam}, \orgaddress{\street{De Boelelaan 1111}, \city{Amsterdam}, \postcode{1081HV}, \state{Noord-Holland}, \country{the Netherlands}}}

\affil[2]{\orgdiv{Mechanical Engineering}, \orgname{École Polytechnique Fédérale de Lausanne}, \orgaddress{\street{LIS-IMT-STI, Room MED1 1126}, \city{EPFL}, \postcode{1015}, \state{Lausanne}, \country{Switzerland}}}

\affil[3]{\orgdiv{Computer Science}, \orgname{New York University Abu Dhabi}, \orgaddress{\street{Saadiyat Island}, \city{Abu Dhabi}, \country{United Arab Emirates}}}

%%==================================%%
%% Sample for unstructured abstract %%
%%==================================%%

\abstract{
In this paper, we introduce Hebbian learning as a novel method for swarm robotics, enabling the automatic emergence of heterogeneity. Hebbian learning presents a biologically inspired form of neural adaptation that solely relies on local information. 
By doing so, we resolve several major challenges for learning heterogeneous control:
1) Hebbian learning removes the complexity of attributing emergent phenomena to single agents through local learning rules, thus circumventing the micro-macro problem; 2) uniform Hebbian learning rules across all swarm members limit the number of parameters needed, mitigating the curse of dimensionality with scaling swarm sizes; and 3) evolving Hebbian learning rules based on swarm-level behaviour minimises the need for extensive prior knowledge typically required for optimising heterogeneous swarms. This work demonstrates that with Hebbian learning heterogeneity naturally emerges, resulting in swarm-level behavioural switching and in significantly improved swarm capabilities. It also demonstrates how the evolution of Hebbian learning rules can be a valid alternative to Multi Agent Reinforcement Learning in standard benchmarking tasks.}

\keywords{Swarm Robotics, Hebbian Learning, Emergence, Heterogeneous swarms}

%%\pacs[JEL Classification]{D8, H51}

%%\pacs[MSC Classification]{35A01, 65L10, 65L12, 65L20, 65L70}

\maketitle
\section{Introduction}
Collective swarming behaviours in nature demonstrate the advantage of effective collaboration in groups of animals \cite{reynolds1987flocks, ramdya2015mechanosensory, puckett2018collective, cavagna2022marginal}. Unfortunately, leveraging such behaviours in robot swarms presents a significant challenge: defining what the individual robots should do for the desired collective behaviour to emerge \cite{hasselmann2021empirical, kegeleirs2025towards}. The challenge arises from the fact that emergent behaviours cannot be defined at the microscopic level, the so called `micro-macro' problem \cite{dorigo2021swarm, BrambillaReview2013}. \textcolor{black}{This problem is exacerbated when considering a heterogeneous swarm, where each individual member may have different behaviours based on different control laws \cite{bayindir2016review, birattari2021automode}, tasks \cite{ferrante2015evolution, ramos2024automatic} or variation in information \cite{raoufi2023, mengers2024leveraging}.} The complexity of individual interactions constrains the field of swarm robotics to heuristic optimisation of simple control strategies in homogeneous swarms \cite{trianni2014evolutionary, scholz2018rotating, sun2023mean, kuckling2023recent, li2024distributed, karaguzel2020collective, rincon2024collective}. \textcolor{black}{For example in some work, copies of the same neural network (NN) are optimised using a black-box Evolutionary Algorithms (EA) \cite{nolfi2016evolutionary, eiben2015Introduction, van2024model, ramos2019evolving, schilling2019learning}. A homogeneous approach can be sufficient at obtaining collective behaviour in which any behavioural specialization can be attributed only to  `phenotype differentiation' (different actions that are a result of different sensorial inputs for the same control laws)}. Heterogeneity in the control laws in the swarm remains a promising frontier, leading to a potentially richer space of swarm behaviours with more pronounced specialisations between members.

\textcolor{black}{Learning-based methods have become a popular approach to automate the design of robot controllers \cite{orr2023multi}. Current state-of-the-art learning methods in the field of Multi-Agent Reinforcement Learning (MARL, \cite{waibel2009genetic, bredeche2022social, bettini2023heterogeneous, albrecht2024multi}) enable specialisation through a centralised learning and decentralised deployment scheme. 
% Here, a centralised evaluator, is trained to teach individual policies (typically a NN)  to control decentralised agents. 
%Here, a centralised evaluator, $Q(\mathbf{x}, a_1 \dots, a_N)$ with $N$ agents,  is trained to teach individual policy networks $\Pi = \{\pi _1, \dots , \pi_N\}$ to control decentralised agents. 
Unfortunately such an approach requires significant training effort as 1) the same micro-macro problem persists (often referred to as the credit-assignment problem \cite{waibel2009genetic, bredeche2022social, bettini2023heterogeneous, albrecht2024multi}) and 2) scalability issues occur when increasing the swarm size (a curse of dimensionality \cite{ orr2023multi}), 
as the joint action and state space can grow exponentially with the number of agents \cite{hao2022breaking}. 
%as the joint action space ($|a_n|$) with the joint state space ($\mathbf{x}$) can grow exponentially with the number of agents ($N$): $\mathcal{O}(\mathbf{x}\cdot|a_n|^N)$ \cite{hao2022breaking}. 
Additionally, decentralised execution in MARL assumes a prior structure after training, making it less useful for swarm application when re-assigning agents to different tasks, or changing swarm size.}

\begin{figure}[ht]
    \centering
    \includegraphics[trim={0 0 7.5cm 0},clip,width=\linewidth]{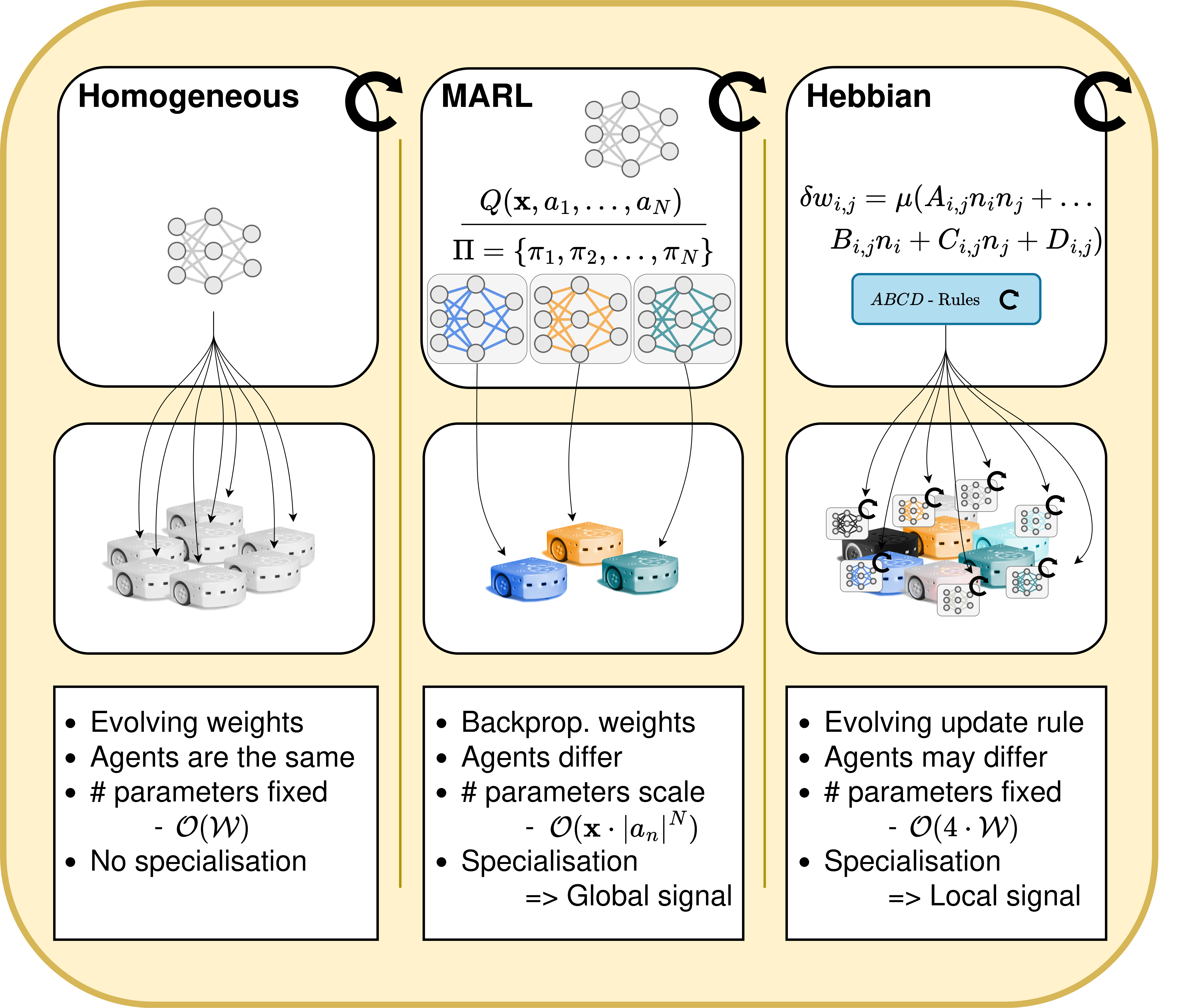}
    \caption{\small 
    Overview of the current methods for obtaining swarm controllers with respect to our Hebbian approach. Note the similarity of Hebbian learning with homogeneous optimisation, as copies of the same $ABCD$-rules are used, without losing the ability to specialise.
    In MARL, a centralised evaluator, $Q(\mathbf{x}, a_1 \dots, a_N)$ with $N$ agents,  is trained to teach individual policy networks $\Pi = \{\pi _1, \dots , \pi_N\}$ to control decentralised agents. \textcolor{black}{Importantly, note how for Hebbian and Homogeneous the number of parameters only depends on the task complexity and not on the number of agents, while in MARL the number of parameters of the centralized evaluator is determined by the size of the joint action space ($|a_n|$) with the joint state space ($\mathbf{x}$), and grows exponentially with the number of agents ($N$): $\mathcal{O}(\mathbf{x}\cdot|a_n|^N)$.} }
    \label{fig:method_overview}
\end{figure}

In this work, we propose a novel approach to learning heterogeneous swarm control by optimising Hebbian learning rules. Hebbian theory, rooted in biology, suggests that synaptic strength between neurons increases through persistent local activity \cite{hebb2005organization}. This principle defines local weight adaptations based on \textit{correlational activity} of pre- and post-synaptic neurons \cite{gerstner2002mathematical}. At the heart of Hebbian learning lies the local update rule ($\delta w_{i,j}$) that relates activities of neighbouring neurons to an update of the connecting weight, see \autoref{eq:abcd}: 
\begin{equation}\label{eq:abcd}
    \delta w_{i,j}=\mu\left(A_{i,j} n_{i}n_{j} + B_{i,j}n_{i} + C_{i,j}n_{j} + D_{i,j}\right)
\end{equation}
\textcolor{black}{Here, for a specific pair of neurons ($i,j$), we update the connecting weight ($w_{i,j}$) based on the neuron activities ($n_i$,$n_j$) and a set of scalar rules ($A_{i,j}$,$B_{i,j}$, $C_{i,j}$, $D_{i,j}$) with a constant learning rate $\mu=0.1$ (not subject to evolution). A more detailed explanation of the Hebbian learning update can be found in \autoref{sec:met_heb}. Note that for this method only a local signal is required (e.g. either sensorial input or activation value of other neurons, with no external teaching), with updates based on the scalar $ABCD$-rules \cite{najarro2020meta, ferigo2022evolving, ferigo2025totipotent}.}

\textcolor{black}{A diagram comparing Hebbian learning with traditional swarm evolution and MARL is illustrated in \autoref{fig:method_overview} (for a more detailed implementation of Hebbian learning we refer to \autoref{sec:meth}). What makes Hebbian learning unique is its indirect optimisation of the agents' control. Optimisation is applied on a meta-learning level (by improving local update rules), which results in a decentralised learning with decentralised control scheme \cite{soltoggio2007evolving,mattiussi2007analog, floreano2008neuroevolution}.
By leveraging Hebbian learning's local adaptivity, we propose to obtain heterogeneity by means of local `unique' experiences that ultimately lead to different NN. 
In our approach, we provide copies of the same $ABCD$-rules to every agent in the swarm and allow different controllers to emerge naturally within the swarm, as individual experiences result in unique updates for each member. Hebbian learning thus requires a fixed number of optimisation parameters (similar to the homogeneous swarm approach) but can achieve heterogeneity through optimisation of individual neural network controllers (similar to MARL but scalable to large number of agents).}

% Optimisation is applied on a meta-learning level, where we focus on improving adaptation, by evolving Hebbian learning rules (using CMA-ES \cite{hansen2001cmaes}). Each robot in the swarm receives the same set of Hebbian learning rules, one rule per weight, denoted as ``ABCD-rules'' ($A,B,C,D \in [-5,5]$), which define the per weight adaptation \cite{najarro2020meta, ferigo2022evolving, ferigo2025totipotent}. Thus, in our case (see \autoref{sec:met_con}), a network with 180 weights ($N_{weights} = 180$) has a total of $4\times N_{weights}=720$ rules. \autoref{eq:abcd} (also shown in \autoref{fig:setup}) presents the update of a single weight ($w_{i,j}$) based on the activities of the pre-synaptic neuron ($n_{i}$) and post-synaptic neuron ($n_j$) connected through $w_{i,j}$. Even though swarm members have the same set of ABCD-rules, their unique local sensory input results in a unique local neuron activity pattern. This dynamic makes the NN weights in the swarm to diverge during deployment, thus heterogeneity emerges. 

The novelty of our approach is in the emergence of heterogeneity through Hebbian learning. By doing so, this method elegantly solves the aforementioned challenges of robot learning for heterogeneous swarms: 1) local updates circumvent the micro-macro problem; 2) Homogeneity of Hebbian rules prevents the curse of dimensionality (fixed number of design variables); 3) by using heuristic optimisation we require minimal prior knowledge (performance is only based on swarm level). We demonstrate that with Hebbian learning heterogeneity indeed naturally emerges, enabling behavioural switching and significantly improving the capability of the swarm.

\section{Results}\label{sec:res}
\subsection{Hebbian learning in simulated MARL benchmarks}
\textcolor{black}{We investigate Hebbian learning in two MARL simulation environments (\texttt{multiwalker\_v9} and \texttt{waterworld\_v4}) provided by Pettingzoo \cite{terry2021pettingzoo}. In the \texttt{multiwalker\_v9} three agents try to walk on a hilly terrain while balancing a beam together. This environment might require specialisation: an agent's position affects its sensor readings and thus require different control strategies for balancing the beam. In the \texttt{waterworld\_v4} environment five agents gather food (that require aggregation of two agents) while avoiding poison. This environment is chosen to shed light on the the scalability of our method as the observation space is large (242 inputs) and the number of agents is higher. We compare our method with the `classic' Homogeneous approach as a Baseline, and two well-established MARL methods provided by the Pettingzoo framework \cite{gupta2017cooperative}: Multi-Agent Deep Deterministic Policy Gradient (MADDPG \cite{lowe2017multi}) and Multi-Agent Twin Delayed Deep Deterministic Policy Gradient (MATD3 \cite{ackermann2019reducing}). Both methods are extensions of single-agent actor-critic algorithms adapted to the multi-agent setting, where agents learn concurrently in a shared environment. A more detailed description of the environmenst and MARL algorithms can be found in \autoref{sec:met_sim}. All experiments were repeated 30 times for statistical analysis.}

\begin{figure}[t]
    \centering
    \includegraphics[height=0.2965\linewidth]{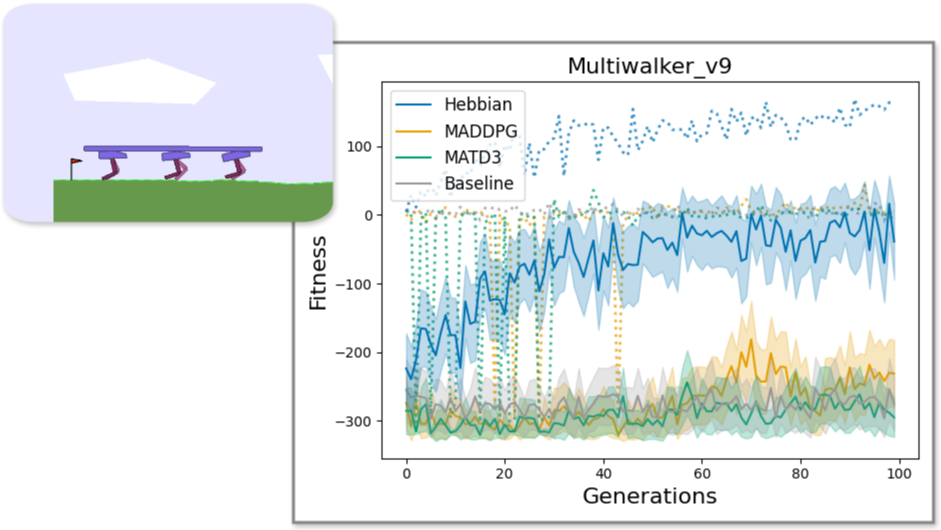}
    % \hspace{1em}
    \includegraphics[height=0.2965\linewidth]{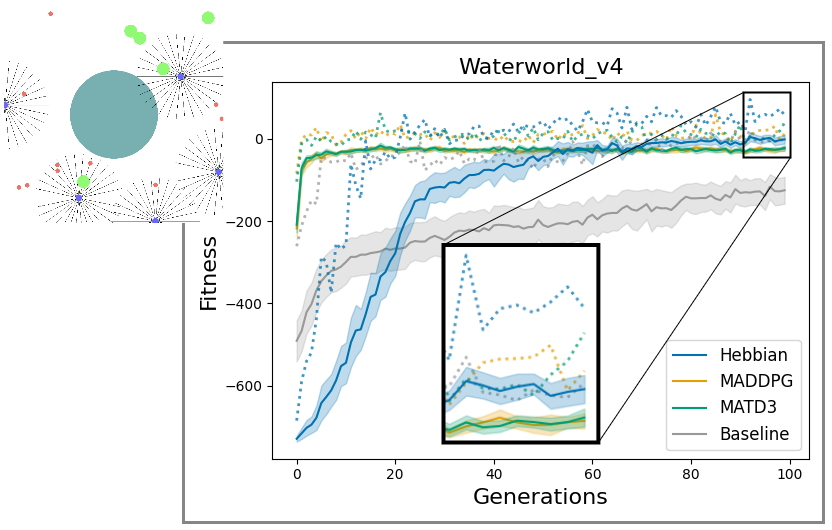}
    \caption{Hebbian learning in multi-agent environments. Solid lines and shaded areas indicate mean ($N=30$) and 95\% confidence interval respectively; dotted lines indicate the best-performing controller. Evolving Hebbian learning rules (Blue) significantly outperforms both MADDPG (Yellow) and MATD3 (Green). The Baseline (Grey, evolving homogeneous control) shows difficulty improving in the \texttt{multiwalker\_v9} environment which indicates that specialisation is beneficial for this task.}
    \label{fig:res_MARL}
\end{figure}

% \begin{table}[]
%     \centering
%     \begin{tabular}{cc}
%     \toprule
%     Hebbian & MADDPG & MATD3 & Baseline
%          &  \\
%          & 
%     \end{tabular}
%     \caption{Caption}
%     \label{tab:my_label}
% \end{table}

\textcolor{black}{The results show that Hebbian learning significantly outperforms both MARL and the Baseline methods, with significantly better performance final performance in either environments. For the \texttt{multiwalker\_v9} $p<0.001$: Hebbian $-38.9\pm55.3$, MADDPG  $-231.4\pm49.24$, MATD3 $-295.4\pm28.1$, Baseline $-275.56.8\pm32.6$; while for the \texttt{waterworld\_v4} $p<0.005$: Hebbian $-0.96\pm10.2$, MADDPG  $-24.0\pm5.62$, MATD3 $-21.6\pm6.57$, Baseline $-124.8\pm32.6$; mean was taken over 30 repetitions and $\pm STD$. For MARL algorithms we found similar results as decribed in . In the \texttt{multiwalker\_v9} we can see that the Baseline rarely improves, which can be explained by the fact that this environment benefits from specialised behaviour (this does not occur in the \texttt{waterworld\_v4} environment). In addition, in environments where more agents are required (\texttt{waterworld\_v4}) both MARL algorithms are unable to improve, indicating that Hebbian Learning is not affected from the curse of dimensionality (see \autoref{tab:sim_env}).}

\subsection{Hebbian learning for simualted and real-robot swarm}
\begin{figure}[hbt!]
    \centering
    \includegraphics[width=\linewidth]{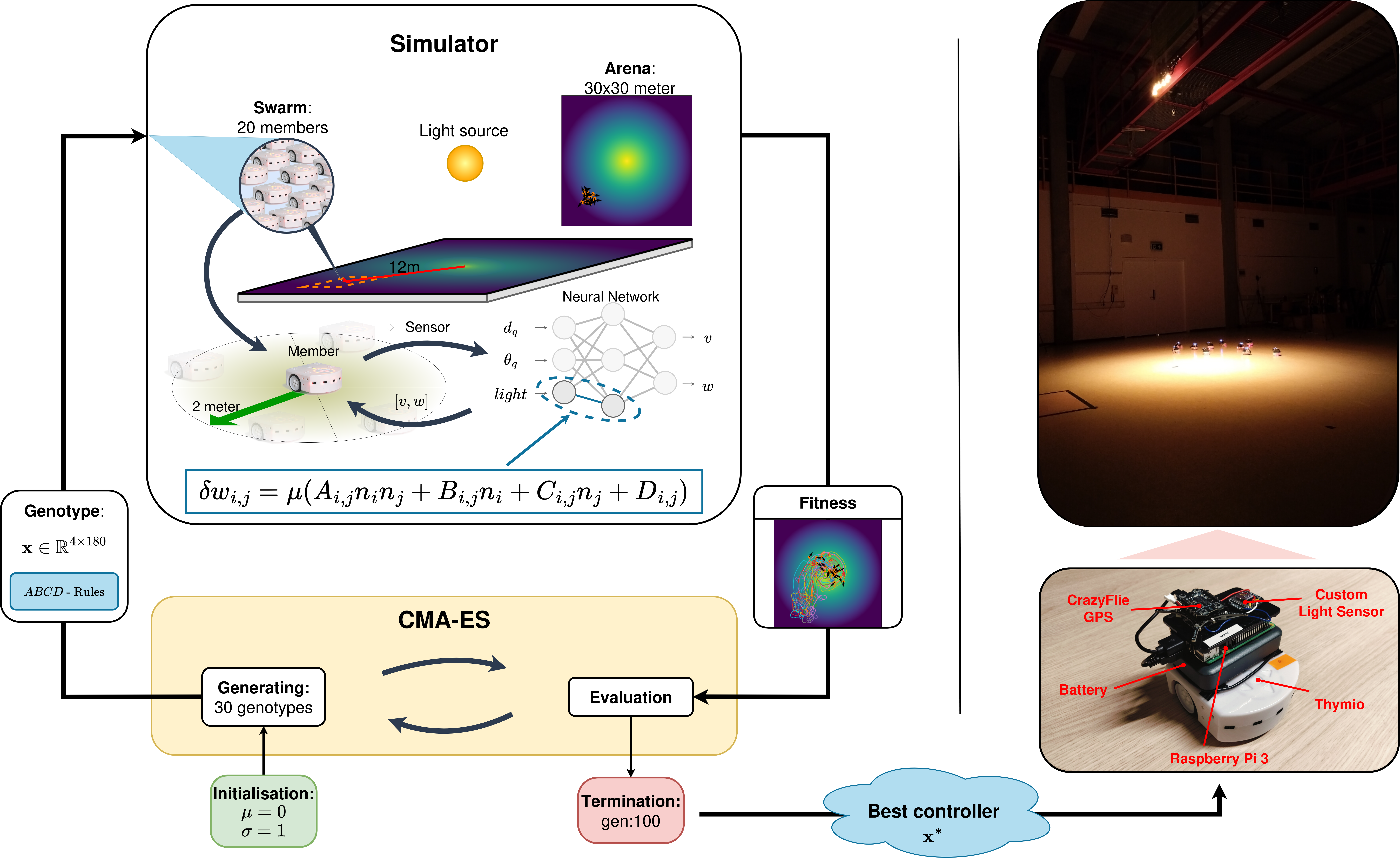}
    \caption{\small 
    Hebbian learning for emergent heterogeneous swarm control for source localisation task. We evolve a set of learning rules (identical between the swarm members) that define an update of the randomly initialised neural networks (different between the swarm members). The resulting set of neural network is unique given the unique experiences between swarm members. After a 100 generations with 30 different Hebbian rules we transfer the best controller from simulation to a real swarm.
    }
    \label{fig:setup}
\end{figure}

\textcolor{black}{We now evolve Hebbian learning rules in simulation for a real-world source localisation task. Furthermore, we analyse the dynamics of the weights under Hebbian learning, and test the \textit{Scalability} and \textit{Flexibility} of the method as additional results in the Appendix. In the end, the overall best controller (over 10 evolutionary experiments with 100 generations with population size 30 in simulation) is transferred to a group of real robots (see \autoref{fig:setup}). Transferring learned controllers from sim-to-real is challenging, and often leads to a drop in performance, also known as the sim-to-real gap \cite{jakobi1995noise}. Mitigating such a drop in performance is often critical \cite{van2021influence,floreano2008evolutionary}, and requires flexibility and scalability in a swarm context. We compare Hebbian learning performance with the homogeneous Baseline method and a scalable Heterogeneous method that trains two specialised sub-groups \textit{Baseline-A} \cite{van2024emergence}. Baseline-A is a heterogeneous control method specifically designed to be more robust, through an online adaptation method (detailed explanation of the methods are presented in \autoref{sec:met_baseline}).}

We evolve Hebbian learning rules using CMA-ES \cite{hansen2001cmaes} in a population of 30 swarms for 100 generations for a source localisation task in simulation (see \autoref{fig:setup} and \autoref{sec:meth} for details of the setup and of the task). The environment used during evolution consists of a single circular gradient with its maximum value in the centre. We use ground robots (Thymio \cite{mondada2017bringing}) with limited capabilities to encourage collaboration (see \autoref{sec:met_rob}). At the start of a single trial, we randomly initialise a swarm of 20 Thymio robots with a neural network consisting of 180 weights in total (uniformly sampled from [-1,1]). \textcolor{black}{The specific network architecture (\autoref{sec:met_con}) is a balance between a large enough network with a small number of learning rules (720 in total), to allows a fair comparison between the \textit{Hebbian} and both Baseline methods, while maintaining robust performance (see Appendix \ref{APP_NAD}). We present additional comparison of different architectures (Appendix \ref{APP:comp_plus}), and corresponding best controller performance in terms of scalability (Appendix \ref{APP:scalability}) and flexibility to different arenas (Appendix \ref{APP:flexibility}).} During deployment, all agents in the swarm update their network weights according to the same Hebbian learning rules, at the same time a new control command is sampled (see \autoref{sec:met_con} for a more detailed explanation). The $ABCD$-rules are tested 3 times (with different random initialisations), with the fitness evaluated on the swarm-level at the end of a trial (see \autoref{eq:fitness}).

\begin{figure*}[ht!]
    \centering
    % \vspace{-1em}
    \begin{minipage}[c]{\textwidth}
        \begin{minipage}[t]{.71\textwidth}
            \subfloat[(a) Best controller in simulation]{\vspace{-1em}
            \subfloat[\small Baseline]{\vspace{-0.5em}\includegraphics[trim={2cm 0 2cm 0},clip, height=0.3\textwidth]{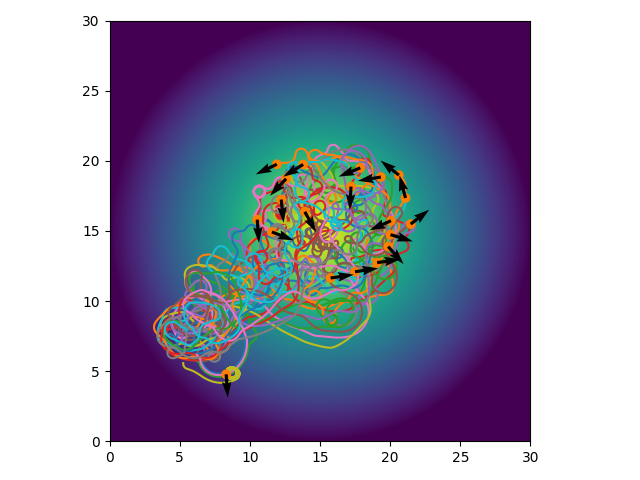}}

            \subfloat[\small Baseline-A]{\vspace{-0.5em}\includegraphics[trim={2cm 0 2cm 0},clip, height=0.3\textwidth]{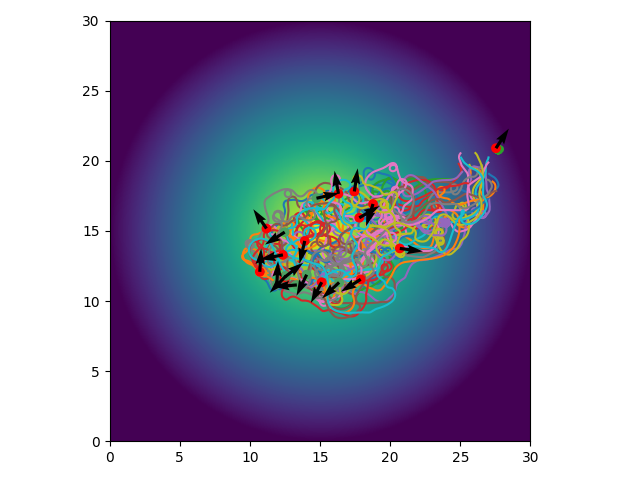}}

            \subfloat[\small Hebbian]{\vspace{-0.5em}\includegraphics[trim={2cm 0 2cm 0},clip, height=0.3\textwidth]{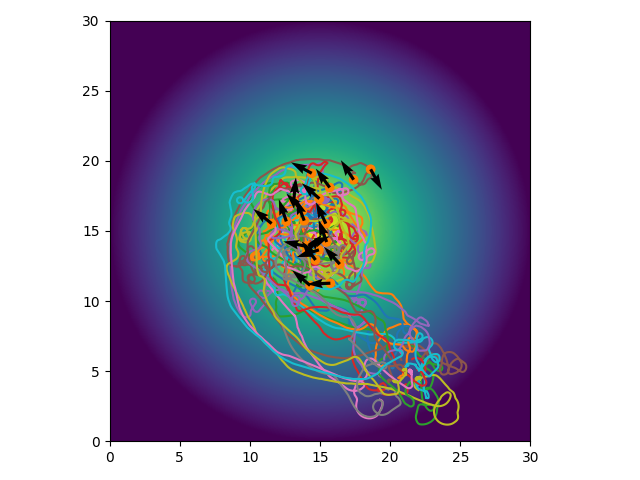}}}
        \end{minipage}
        % \hfill
        \hspace{-2.5em}
        \begin{minipage}[t]{.28\textwidth}
            \subfloat[(b) Learning curves]{\vspace{-0.5em}\includegraphics[trim={1.05em 0 1.05em 0}, clip,height=0.9\textwidth]{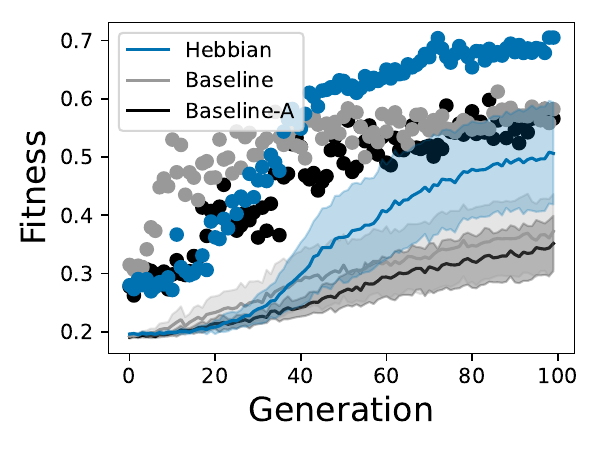}}
        \end{minipage}
\end{minipage}
    \caption{\textbf{a)} Final behaviour of the overall best swarm per method: Baseline (left), Baseline-A (middle), Hebbian (right); \textbf{b)} Learning curves over generations. Lines indicate mean values with shaded area providing 95\% confidence bounds $N_{runs}=10$, dots indicate maximum value per generation. Hebbian learning (Blue) vs. Baseline (Orange).}
    \label{fig:Results}
\vspace{-1em}
\end{figure*}

Similar to the previous comparison we find that Hebbian learning outperforms the other methods, in \autoref{fig:Results}. This is visible in the progression of the fitness curves (\autoref{fig:Results}b), as well as the behaviour of the best overall controllers (\autoref{fig:Results}a). A visual inspection of these overall best controllers is made to analyse the emergent swarm behaviour. The Baseline controller induces emergent gradient sensing behaviour by a circular flocking pattern that is persistent throughout the trial (\autoref{fig:Results}a, left). A video of this trial can be found at \url{https://youtu.be/foHiA5m4o1M}. \textcolor{black}{In the video of the \textit{Baseline-A} controller we initially observe a similar circular swarming pattern emerge (when the swarm is majority green) that changes at the centre when more swarm members adapt their behaviour toward scattered random aggregation (majority red) \url{https://youtu.be/aPJ8Dq-imLo}.} For \textit{Hebbian learning}, online behavioural switching occurs during deployment (\autoref{fig:Results}a, right). At the start, on the outer-section of the arena, the swarm scatters in a circular pattern to assess gradient information (similar to the Baseline). Once the gradient is found, the swarm switches collective behaviour and collectively moves following a (almost) straight-line towards the centre. This behaviour emerges around 6 seconds into this video: \url{https://youtu.be/Rb3K-9-z9II}. Arriving at the centre, the swarm loses its formation and slowly aggregates at high light intensities. The difference between the Baseline and Hebbian swarm behaviour indicates that online adaptability successfully encodes multiple behaviours that lead to more efficacy (i.e., maximum performance).

\begin{figure}[bh!]
\centering
\begin{minipage}[c]{0.95\linewidth}
    \begin{minipage}[t]{.3\linewidth}
     \begin{minipage}[t]{0.96\linewidth}
        \centering
        \subfloat[(a) Static environment]{\textcolor{black}{\fbox{\includegraphics[trim={8em 2.75em 8em 1em},clip,width=\textwidth]{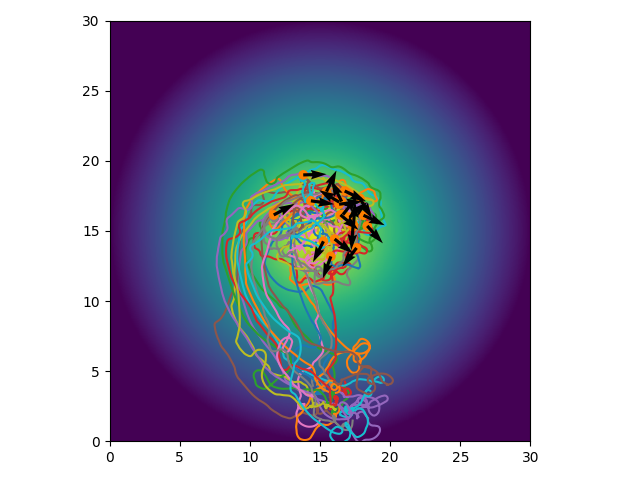}}}}
      \end{minipage}
     \end{minipage}
     \hspace{0.5cm}
    \begin{minipage}[t]{.6\linewidth}
        \centering
        % Subfigure (a) with two images inside
        
        \subfloat[(b) Dynamic environment (switch at $t=300$)]{\textcolor{black}{\fbox{
            \begin{minipage}[t]{0.48\linewidth} 
                \centering
                \includegraphics[trim={8em 2.75em 8em 1em},clip,width=\textwidth]{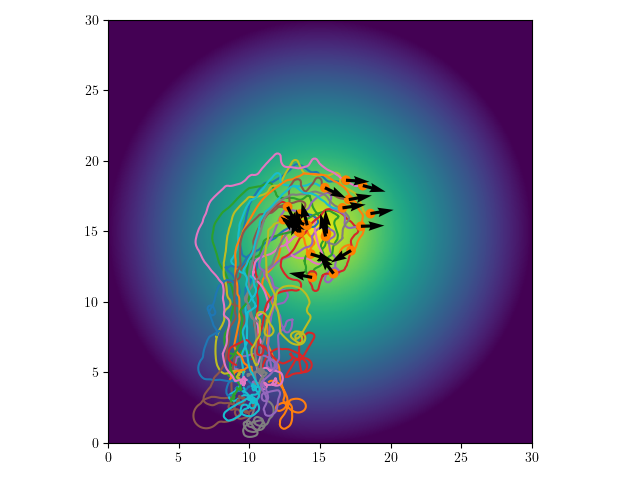}
            \end{minipage}
            % \hfill
            \begin{minipage}[t]{0.48\linewidth} 
                \centering
                \includegraphics[trim={8em 2.75em 8em 1em},clip,width=\textwidth]{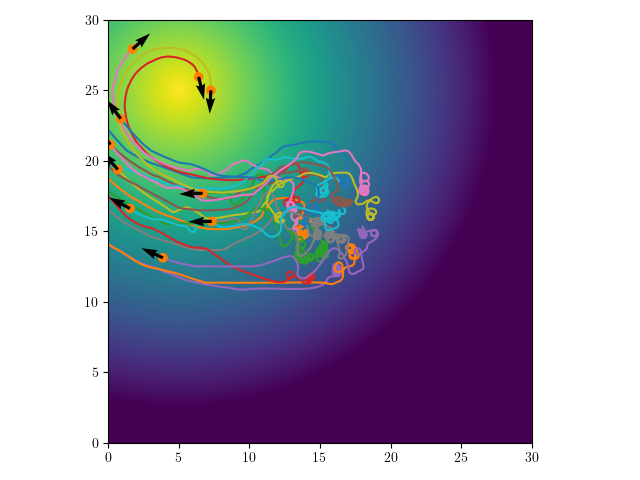}
            \end{minipage}
            % \hfill
        }}}
    \end{minipage}
     % \hspace{0.5cm}
\end{minipage}
\begin{minipage}[c]{0.99\linewidth}
\centering
        \begin{minipage}[t]{.32\linewidth}%## ROW 10x10
            \centering
            \subfloat[(c) Swarm light intensity]{\includegraphics[trim={1em 0em 1em 1em},clip, width=\textwidth]{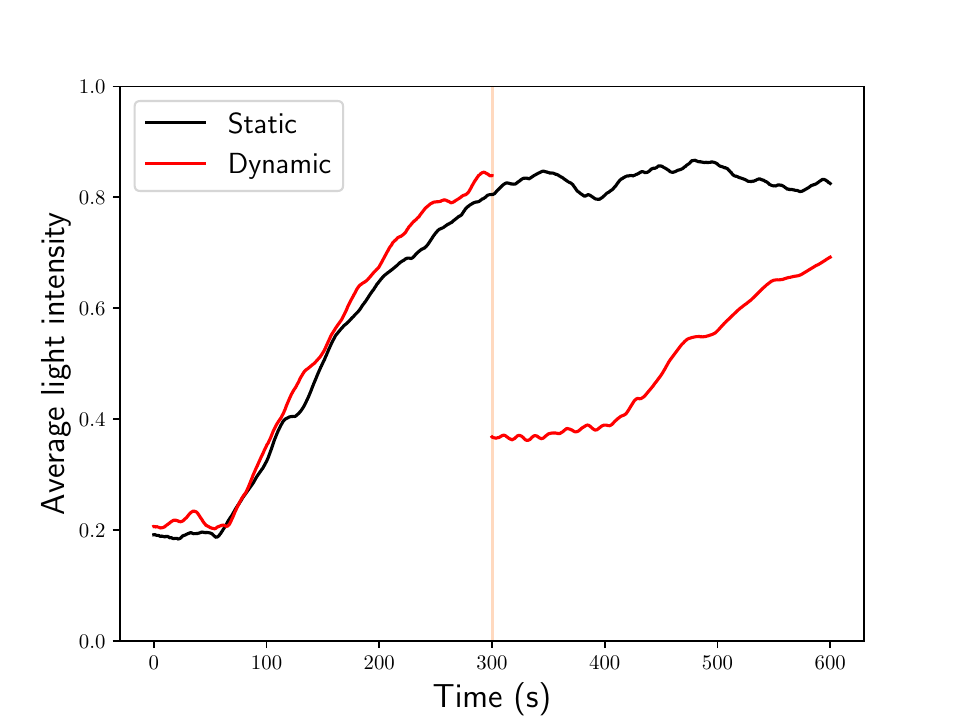}}
        \end{minipage}
% \hspace{2em}
        \begin{minipage}[t]{.32\linewidth}
            \centering
            \subfloat[(d) Weight autocorrelation]{\includegraphics[trim={1em 0em 1em 1em},clip,width=\textwidth]{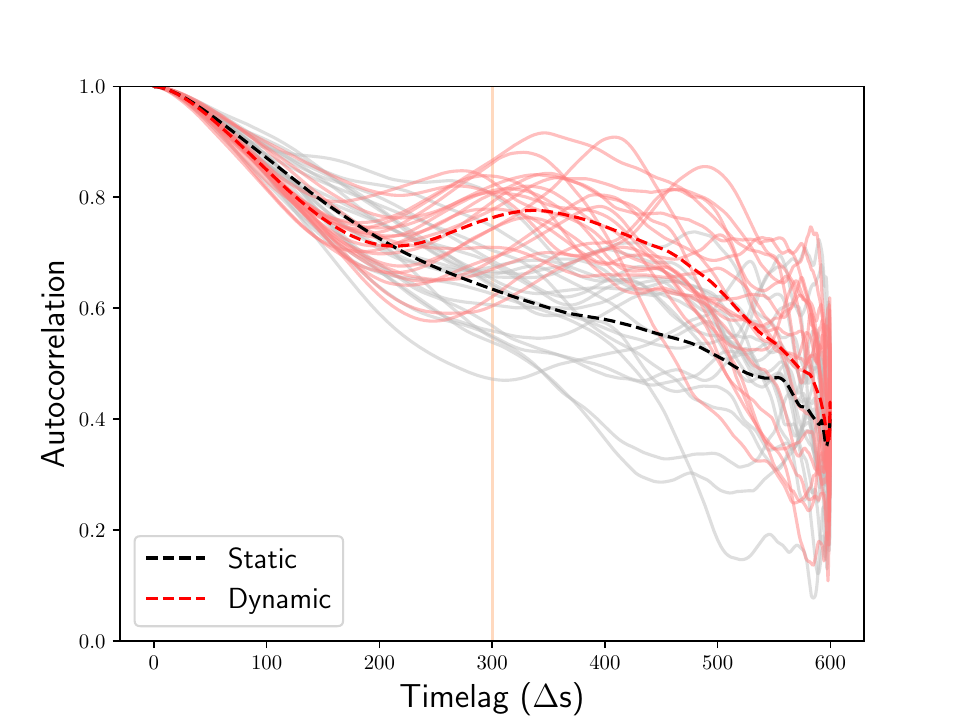}}
    \end{minipage}
% \hspace{2em}
        \begin{minipage}[t]{.32\linewidth}
            \centering
            \subfloat[(e) Weight variability]{\includegraphics[trim={1em 0em 1em 1em},clip, ,width=\textwidth]{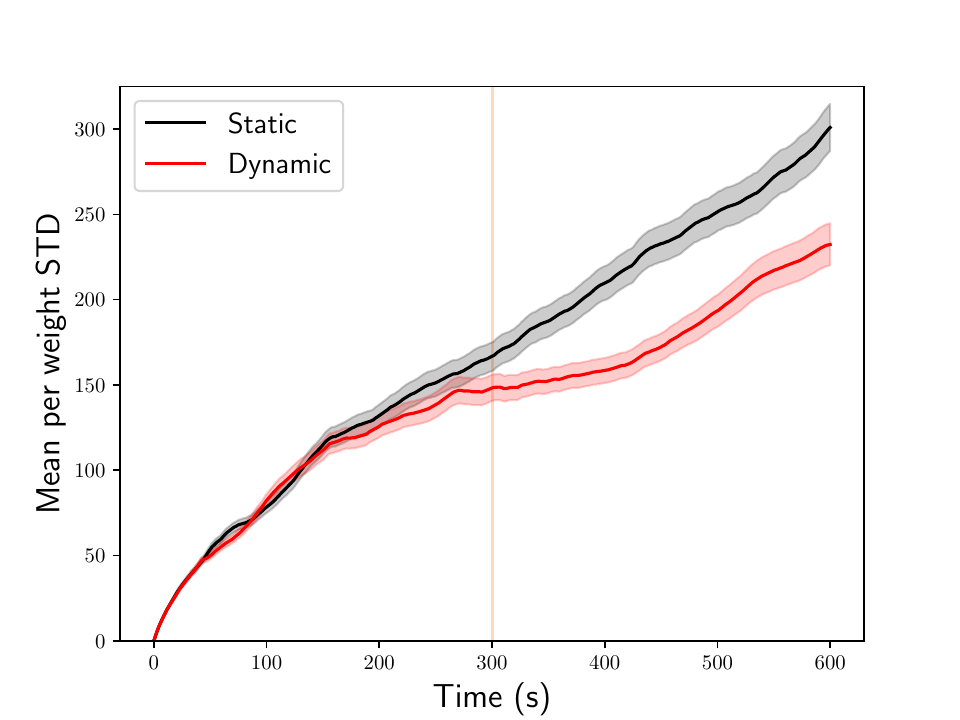}}
    \end{minipage}
\end{minipage}
\caption{\small Retesting best Hebbian controller in two conditions: static (black, a) dynamic (red, b) where the light source abruptly changes position halfway through the trial (indicated as yellow vertical lines in c-e). We analyse: (c) average light intensity over time; (d) weight autocorrelation within members; (e) weight variability between swarm members. }
\label{fig:dyn_v_stat}
% \vspace{-1em}
\end{figure}

\textit{\\Hebbian weight dynamics}\\
We present the weight dynamics of the best Hebbian controller during deployment (\autoref{fig:dyn_v_stat}). \textcolor{black}{For clarity of the trajectories, a single run is presented but the results are consistent over many repetitions} The weight dynamics provide insight into the heterogeneity of our swarm and if adaptation continuously persists throughout the run or if the NN network converge towards the same behaviour. To this end, weight values are logged in two experimental conditions: 1) a static environment which is identical to the task during evolution ($30\times30$ metre arena with circular gradient for 600 seconds); 2) a dynamic environment where the position of the circular gradient is perturbed halfway through the run (at t=300 seconds).\label{sec:dyn_vs_stat}

\autoref{fig:dyn_v_stat} shows the results of both experiments, with the light source perturbation condition visible in \autoref{fig:dyn_v_stat}b (videos are available for the \textit{static}: \url{https://youtu.be/B1KynO2KL_Q}); and \textit{dynamic}: \url{https://youtu.be/ykdg02hDXzs}. \textcolor{black}{Both static and the first half of the dynamic conditions (\autoref{fig:dyn_v_stat}a; \autoref{fig:dyn_v_stat}b left) show the same behavioural switching during deployment, i.e., circular scattering at t=0 followed by straight line collective movements towards the centre. In the second half of the dynamic condition (\autoref{fig:dyn_v_stat}b right) this behavioural switch reoccurs after the light source is perturbed. These results indicate that the dynamic online switching capability imposed by the Hebbian rules is not only a transient property, but a persistent adaptive capability imposed by the Hebbian rules throughout the run.}

We find more support of the persistent presence of swarm-level behavioural switching, by analysing the dynamics of the weights. Comparison of the weight dynamics during deployment of the dynamic and static condition (\autoref{fig:dyn_v_stat}c-e), reveals that the weights are undergoing a continuous adaptation process that does not converge. Here, the vertical orange line in \autoref{fig:dyn_v_stat}c-e indicate the time where we perturb the light source for the dynamic condition. 

\textit{\\Average light intensity}\\
\autoref{fig:dyn_v_stat}c shows the average light intensity of all swarm members over time (see \autoref{eq:fitness_t} for the full calculation). In the first half of the experiment, both conditions show a similar increase in light intensity. At the time of the switch average light intensity drops, reflecting the change in the dynamic environment. Notably, all members start to rotate at their spot at the same frequency and phase, visible in both \autoref{fig:dyn_v_stat}b right (small trajectory circles) and c (average fitness going up and down). We suspect the low light intensity to clip the NN output towards this behaviour, while the weights re-adapt. Around $\sim$100 seconds after the perturbation, the swarm shows the ability to sense the new gradient and switches back towards a more targeted collective movement. 

\textit{\\Average mean weight autocorrelation}\\
The average autocorrelation indicates how weight dynamics are similar between different time-lags during deployment (\autoref{fig:dyn_v_stat}d), see \autoref{eq:autocor} for the full calculation. The increase in autocorrelation supports the hypothesis that the swarm readapts to initial behaviours required to sense and follow the gradient. \textcolor{black}{The rise in autocorrelation prior to the perturbation is a natural consequence of the autocorrelation computation, which compares weights over a time window. Here, the correlative similarity \textit{increase} until the time-lag equals a full period length at maximum correlation at $\sim$350 seconds. This indicates that 1) adaptation occurs gradually after the abrupt switch (it takes $\sim50$ seconds to revert to the initial NN dynamics; 2) The weights do not fully revert to their initial values, which makes sense as initially weights were randomly sampled at $t=0$.} 

\textit{\\Average mean weight STD}\\
The average weight variance (defined as the mean $std$ of the weights) measures the amount divergence between agents` weights (\autoref{fig:dyn_v_stat}e), see \autoref{eq:meanstd} for the full calculation. \textcolor{black}{If each robot's network is the same --Note that the weights inside the network can differ-- then the value of this metric is 0}. The mean weight STD slowly increases over time, indicating a higher variance between NN controllers thus increase in heterogeneity. Interestingly, even though heterogeneity between agents kept increasing collective behaviour was maintained. Closer inspection of the weights (see Appendix \ref{APP_hists}) shows that during the run weights monotonically diverge (we did not limit the size of the weights). After the light perturbation, some of these weights revert to lower values \textcolor{black}{(see higher concentration of weights around 0 in Appendix \ref{APP_hists} t450). The shows as a plateauing of the mean STD in \autoref{fig:dyn_v_stat}e}. After the new gradient is found, weights start to monotonically increase again.

\begin{figure*}[ht!]
    \centering
    % \vspace{-1em}
    \begin{minipage}[c]{\textwidth}
    \centering
        \begin{minipage}[t]{.71\textwidth}
            
            \subfloat[(a) Best controller in the real world]{\vspace{-1em}
            \subfloat[\small Baseline]{\vspace{-0.5em}\includegraphics[trim={4em 1.2em 11.0em 3.6em}, clip, height=0.3\textwidth]{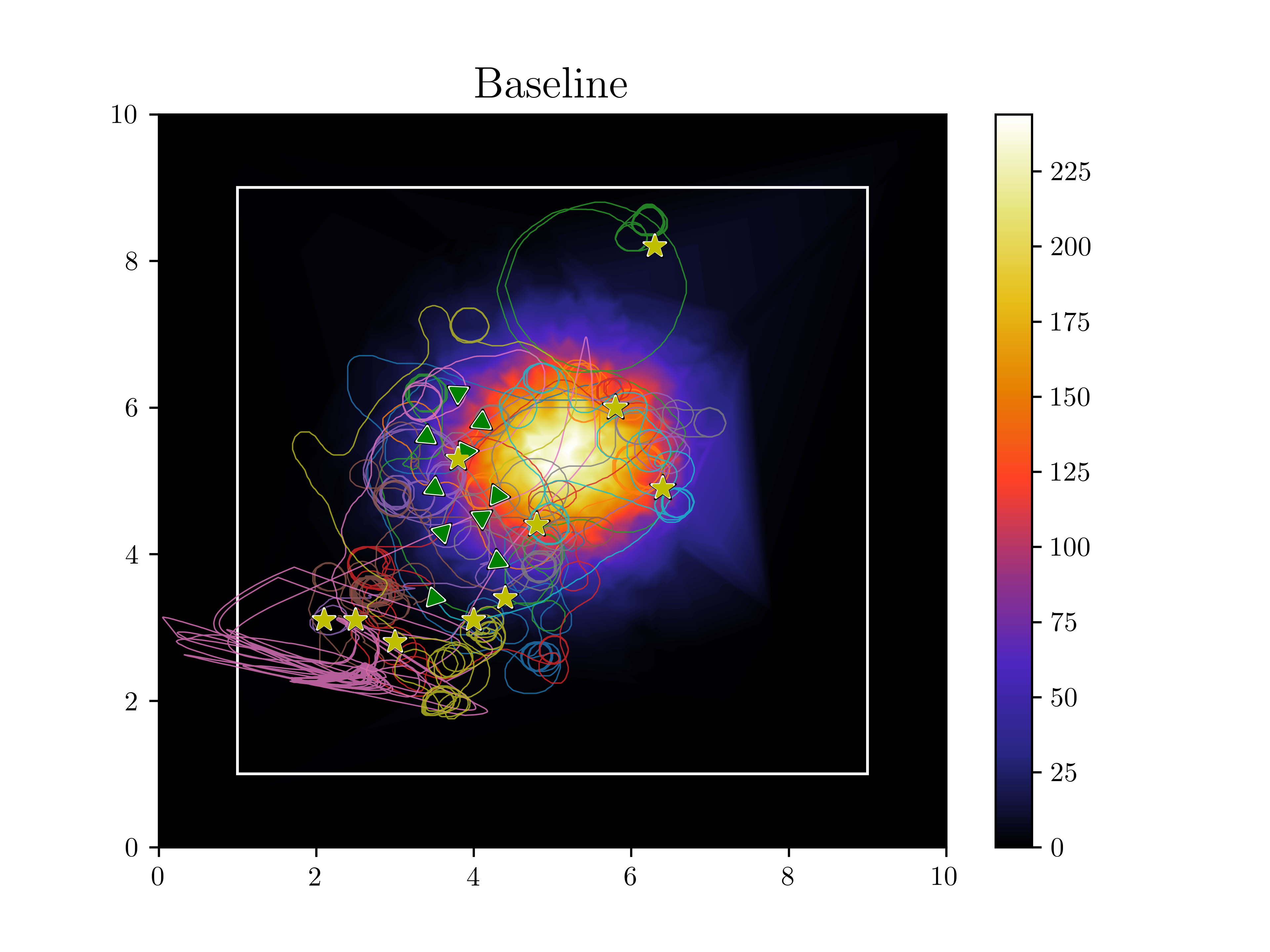}}

            \subfloat[\small Baseline-A]{\vspace{-0.5em}\includegraphics[trim={5.1em 1.2em 11.0em 3.6em}, clip,height=0.3\textwidth]{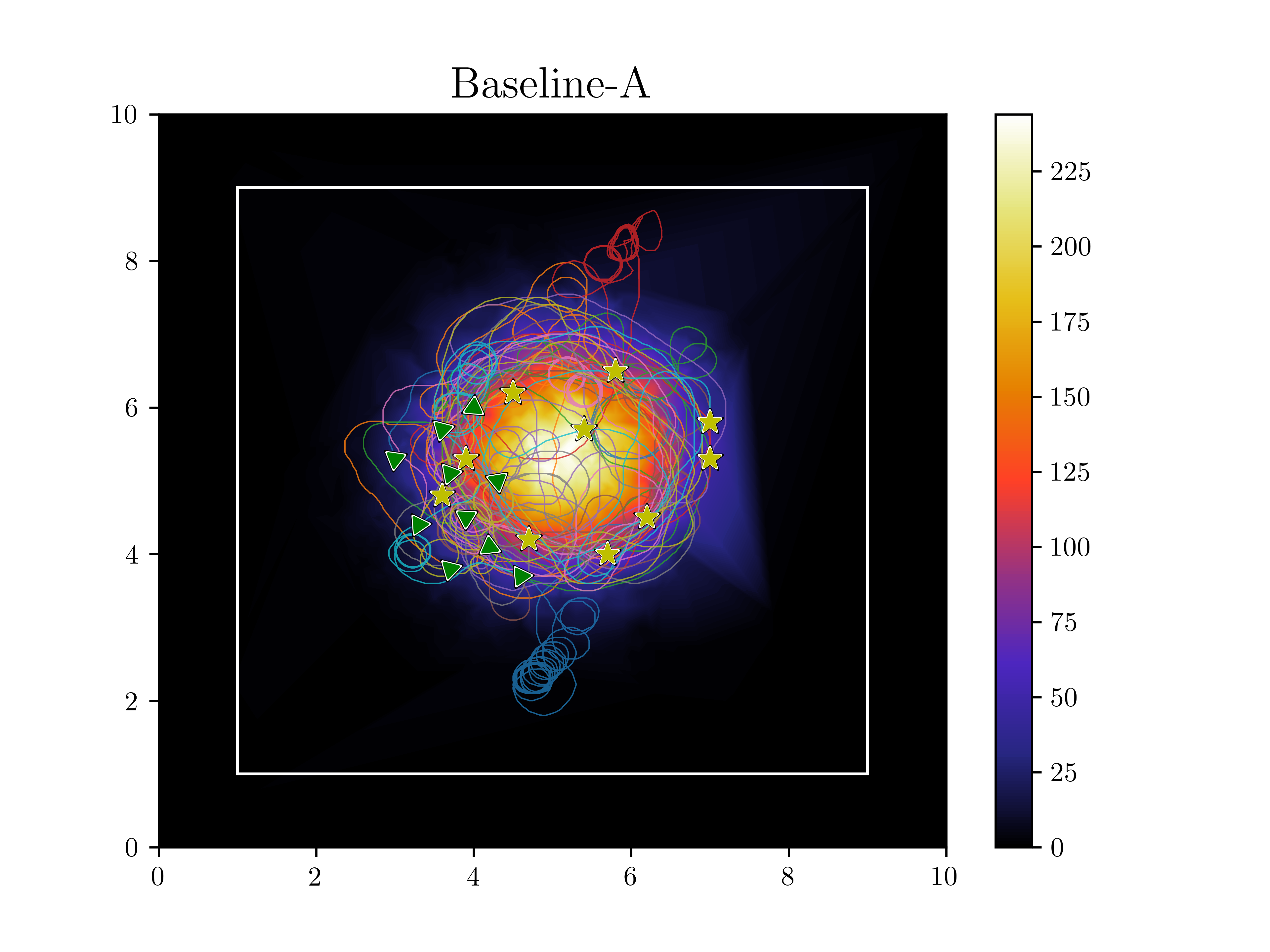}}

            \subfloat[\small Hebbian]{\vspace{-0.5em}\includegraphics[trim={5.1em 1.2em 6.5em 3.6em}, clip,height=0.3\textwidth]{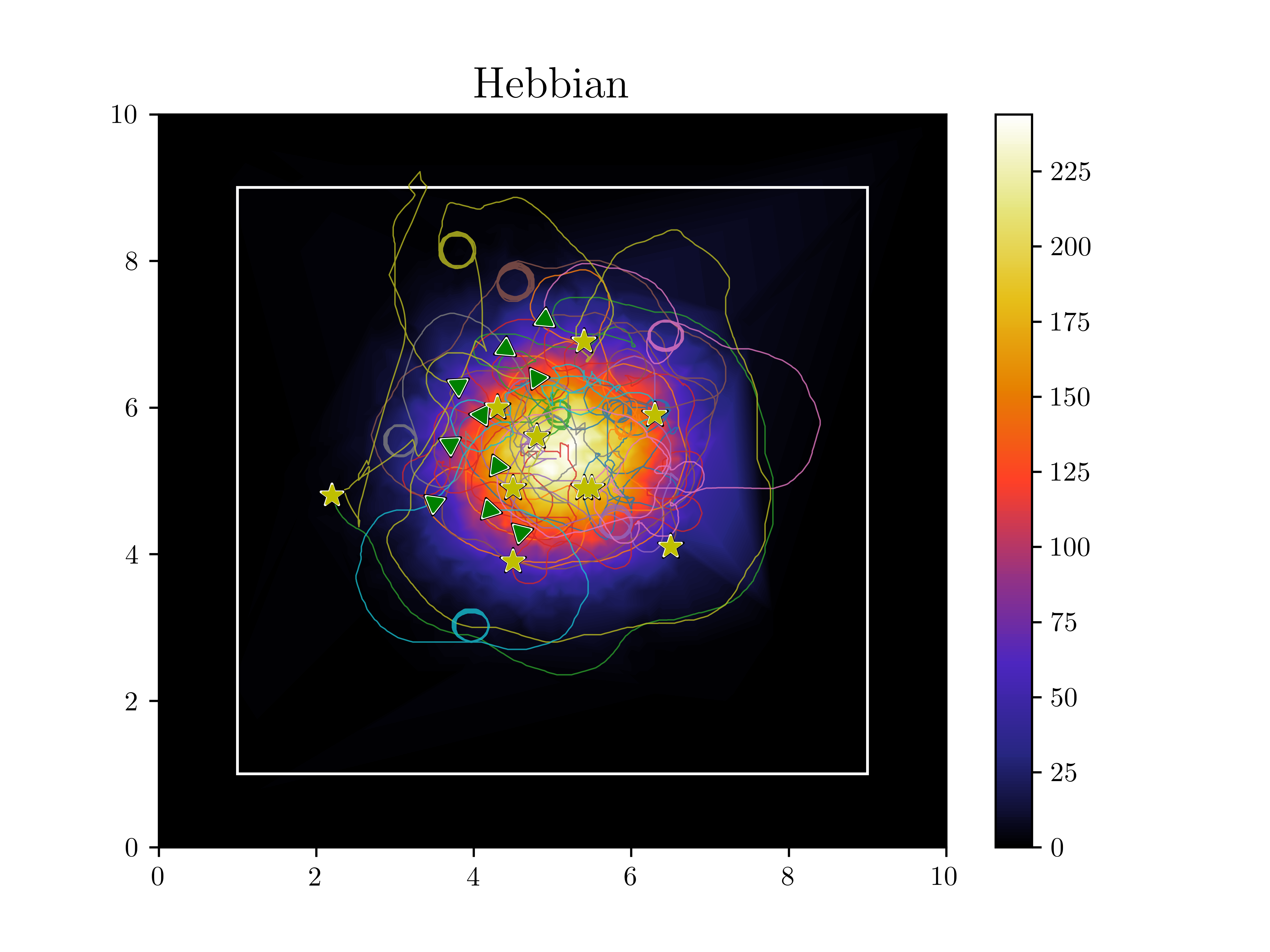}}}
        \end{minipage}
        % \hfill
        \begin{minipage}[t]{.28\textwidth}
            \centering
            \subfloat[(b) Reality gap]{\includegraphics[trim={3.85em 0 4.5em 0}, clip,height=0.95\textwidth]{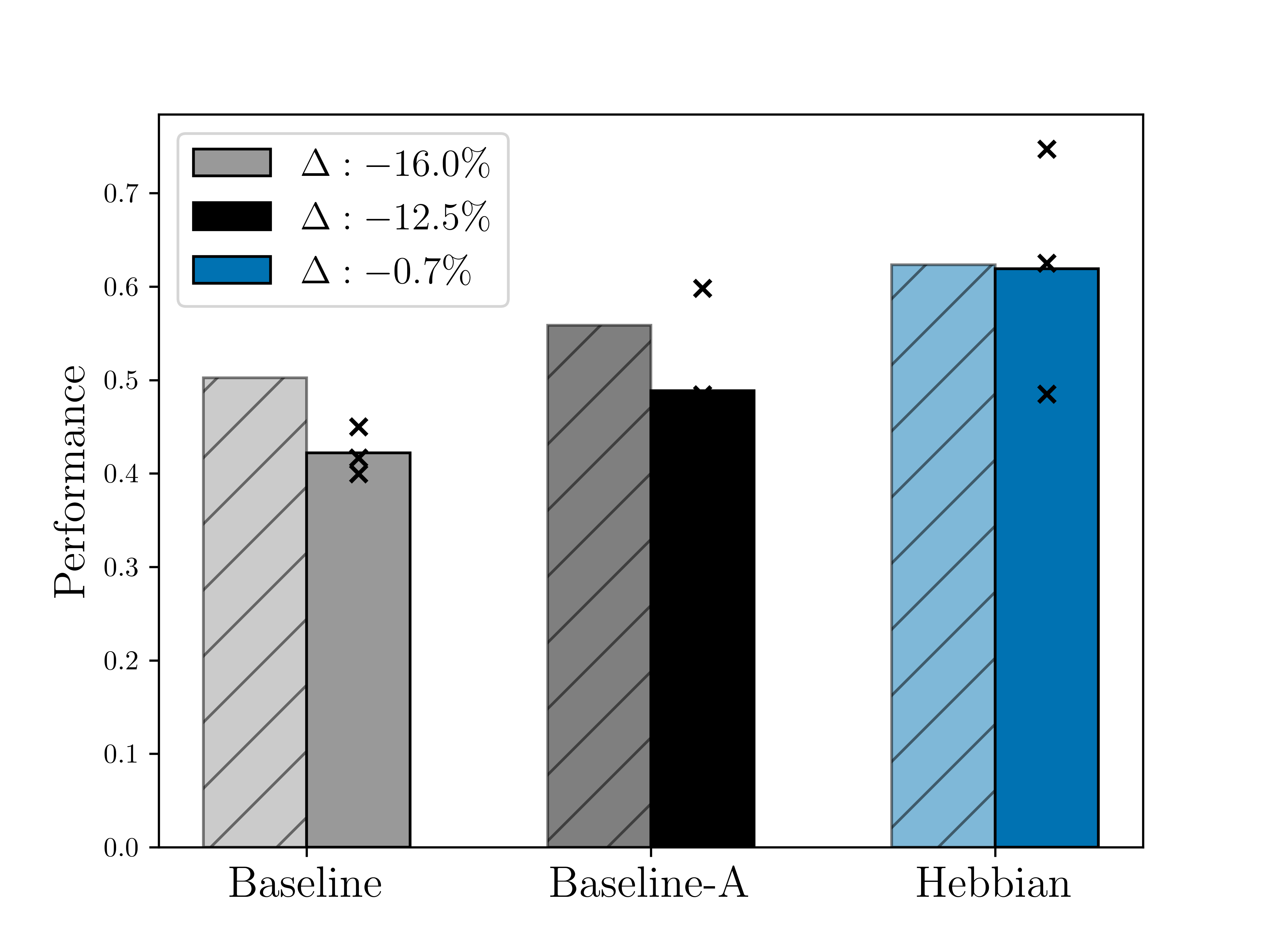}}
        \end{minipage}
\end{minipage}
\caption{\label{fig:res_real} Real world source localisation task. (a) Trajectories of the real world Thymio swarm in a $8\times8$ metre arena (white rectangle), tested for 10 minutes starting at the edge of the gradient (green triangles; end position as yellow stars). (b) Estimated reality gap per condition, with the mean performance in simulation (light coloured bars with stripes, 60 repeated $N=60$); and the mean performance in the real world experiment (dark colored bars, 3 repititions).}
\vspace{-1em}
\end{figure*}

\textit{\\Reality gap}\\
The reality gap in robotics is a term that encompasses the difference between simulated robot behaviour and the actual real-world behaviour \cite{jakobi1995noise}. This difference is often detrimental in automated design, as the optimiser tends to overfit on an unrealistic simulator resulting in a `simulated optimal behaviour' that drops in performance after real world transfer \cite{van2021influence,floreano2008evolutionary}. 
We now transfer our best controllers to a real robot swarm. The scalability and flexibility experiments indicate the viability of both Baseline methods as a benchmark Appendix \ref{APP:scalability}\&\ref{APP:flexibility}. Due to space limitations the resulting swarm was employed in a $8\times8$ metre arena using 10 custom Thymios (see \autoref{sec:met_real}). We show the trajectories of the best runs per controller in \autoref{fig:res_real}a (3 repetitions per controller were done). In all runs, the swarm spread out from the initial positions and moved towards the centre. Similar to the simulated behaviours, we can see both Baseline trajectories engaging a clockwise circular pattern around their centroid (especially Baseline-A the pattern in the middle panel). In the Hebbian learning conditions the real behaviour did not display clear behavioural switching, as the light source was too close to the initial state of the swarm.  

In \autoref{fig:res_real}b, we estimate the reality gap by comparing the real world performance with the simulated swarm (the simulator is adapted to match the real environment accordingly and tested for 60 repetitions). The average the final fitness of all runs provides a rough estimate of difference in performance between the simulated and real environment. Here we see a tendency for both Baseline methods to substantially drop in performance (more than 10 percent) while the Hebbian approach maintains similar average fitness (-0.7\%). Given the previous results on \textit{Scalability} and \textit{Flexibility} this result provides additional support for our findings that Hebbian learning improves consistency of performance when employed in untrained conditions.

\section{Discussion}\label{sec:disc}
The central idea of our method is to efficiently optimise update rules for learning agents, from which heterogeneity can automatically emerge. Hebbian learning addresses three major challenges in current heterogeneous swarm literature simultaneously \cite{bayindir2016review,hasselmann2021empirical, dorigo2021swarm, bettini2023heterogeneous}: 1) online adaptation does not require global information thus circumventing the micro-macro problem; 2) Every agent receives the same set of Hebbian rules, thus solving the curse of dimensionality; 3) minimal prior knowledge allows for more flexibility, \textcolor{black}{as we do not define heterogeneous specialisation beforehand, and can scale swarm size after training}. 
Designing swarm control is a challenging task due to the obfuscation of emergent collective behaviour. Heterogeneous swarm control requires additional care, given the credit assignment problem \cite{dorigo2021swarm}, in-depth know-how on task specialisations \cite{birattari2021automode}, and scaling limitations \cite{albrecht2024multi}.Hebbian learning provides an elegant solution to all these concerns, by optimising local adaptation through a fixed number of ABCD-rules (thus reducing complexity to $4\times N_{weights}$), and allowing heterogeneity to emerge without a `teacher signal'. These advantages were found to significantly improve performance of Hebbian learning with respect to homogenous control and state-of-the-art MARL (\autoref{fig:res_MARL}). This work is the first to apply Hebbian learning for emergent heterogeneous swarm control, and successfully shows a validation in a real-world swarm (\autoref{fig:res_real}).

\textcolor{black}{The online adaptations induced by Hebbian learning represent a generalised mathematical prior of optimal input correlations for a given task during deployment (\autoref{eq:abcd}). While Hebb's rule formulation itself is task-agnostic, practically we observe algorithmic biases in the neural network behaviours. For instance, a tendency for weights to grow over time is evident (Appendix \ref{APP_hists}). Despite this, Hebbian learning consistently outperforms across different tasks, indicating broad applicability and outperformance, regardless of such biases.
Even in a disadvantaged setting of \texttt{waterworld\_v4} (here applying high forces are penalised) Hebbian learning initially underperforms but eventually surpasses benchmark algorithms, further showcasing its robustness and general applicability.
}

Besides an increase in efficacy (\autoref{fig:res_MARL} and \autoref{fig:Results}), we found that Hebbian learning transfers better to the real world (\autoref{fig:res_real}). Based on the Scalability and Flexibility (presented in the Appendix \ref{APP:scalability}\&\ref{APP:flexibility}) this is not surprising, as Hebbian learning is able to perform more consistently than the other methods when changing environments and scaling of swarm size. Interesting to note is the negative correlation with scalability for Hebbian learning. We suspect Hebbian learning is able to consistently find the direction of the gradient, but information propagation may slow with increasing swarm size, resulting in slower collective decision making.

% This indicate the strength of Hebbian learning to adapt to new environments. In addition to the continuous adaptation, we suspect the other methods to possibly suffer from overfitting on the training problem. Hebbian learning naturally induces dynamic interactions between the swarm members, where the reaction between agents that sense eachother changes over time. This induces an additional layer of uncertainty, which prevents overfitting to the trained environment. 

The central concept for obtaining heterogeneity is that local Hebbian adaptations exploit each individual's unique experiences. \textcolor{black}{It should be noted, that this method does not guarantee heterogeneity to emerge. Learning rules could drive neural networks toward a robust attractor in weight space, yielding near-homogeneous control post-initialisation. Additionally, Hebbian learning might transiently exploit random initialisation to locate gradients during early adaptation.
Weight analysis revealed that heterogeneity did emerge during deployment (\autoref{fig:dyn_v_stat} and Appendix \ref{APP_hists}), with swarms re-expressing prior behaviours after light-source perturbation.} Interestingly, continuously growing weights did not functionally impair the ability for behavioural reversion of the swarm. We suspect that neurons with increasing weights become functionally obsolete for task-solving, while those retaining lower-magnitude dynamic connections play a pivotal role in the network's functionality. In the future, we should investigate normalisation techniques for longer deployment time and computational stability \cite{oja1997nonlinear}. 

The behavioural switching found in \autoref{fig:Results} shows that Hebbian learning evolved a strategy of encoding different collective behaviours. For our work, behavioural switching on a swarm-level came as an additional advantage, as the main goal was the emergence of heterogeneity or specialisation inside the swarm. We found behavioural switching consistently emerge over all 10 evolutionary runs, but did not find such a behaviour for the Baseline (previous work always found similar circular patterns \cite{van2022environment} \cite{van2024emergence}). We suspect that the Hebbian learning is more capable of encode multiple behaviours, even when scaling the Baseline to large networks (Appendix \ref{APP_NAD}).

% In addition to behavioural switching Hebbian learning proved to be more capable to cope with unseen environments (\autoref{tab:res_val} and \autoref{fig:res_real}). Online adaptation is highly suited for such a task but can also introduce more chaos into the dynamical system. Luckily, here it proved to add additional robustness to the control strategy. This provides a promising direction for applying Hebbian learning in real robot(s) (swarms).

\textcolor{black}{The source-localisation task showcases outperformance through behavioural switching, Scaling, Flexibility and adapting to the sim-to-real gap. These properties hint at the robustness of Hebbian learning in the context of robot swarms and its generalisability to more complex, practically valuable tasks in the future.} Other applications of Hebbian learning in robots also found similar capabilities, on task-switching \cite{urzelai2001evolution}; meta-learning with fast adaptations \cite{najarro2020meta}; bridging the sim-to-real gap \cite{leung2025bio}, and inducing specialised networks for control \cite{ferigo2025totipotent}. The ability of Hebbian learning to continuously specialise and encode multiple behaviours encourages further investigation. Interesting applications could be presented in continual learning, or adaptive control, robust modelling, and online fine-tuning during deployment. For swarms specifically, Hebbian learning suggest a unexplored frontier of distributed control algorithms with decentralised adaptations. 
On a final note, Hebbian learning provides an interesting case of biologically plausible learning with an embodied intelligence setting \cite{miconi2025neural}. Uncovering optimal mechanisms of learning without backpropagation might shed some light on how biological systems operate. \textcolor{black}{In the future, we will pursue the design of such sophisticated methods to solve more complex tasks in robot swarms, and explore different optimisation techniques for Hebbian learning itself.}

\section{Methodology}\label{sec:meth}
\subsection{MARL experiment design}\label{sec:met_sim}
\textcolor{black}{We initially compare performance of Hebbian learning on two simulation environments (\texttt{multiwalker\_v9} and \texttt{waterworld\_v4} provided by Pettingzoo with default setting as of 24-06-2025 for both the environments and the MARL benchmarks available \cite{terry2021pettingzoo}, see \autoref{tab:sim_env}. We allow a simulation budget of 5.4M timesteps for each environment (corresponding with the total number of timesteps for the Real Hebbian experiment \autoref{sec:met_swarm_exp}). The size of the NN for the Hebbian and Baseline controllers are the same; and is determined by the size of the MADDPG actor policy, such that the number of optimisation parameters for the actor is approximately equal to the Hebbian condition (i.e., the number of learning rules). When evolving the NN controllers (both Hebbian and Baseline) we accumulate rewards until the end of the trial as a final fitness. In MARL experiments, intermediate states, actions and rewards are used to optimise with back-propagation. For both controllers, we follow the default architectural settings that were optimised for each environment \cite{terry2021pettingzoo}.}

\begin{table}[h]
    \caption{Parameters of simulated environment}
    \label{tab:sim_env}
    \centering
    \begin{tabular}{l c c}
         \toprule
       Parameter  & \texttt{multiwalker\_v9} & \texttt{waterworld\_v4}  \\
         \midrule
      Nr. of agents & 3 & 5 \\
      Action space & $\left(-1, 1\right)^{4}$ & $\left[-0.01, 0.01\right]$\\
      Observation space & $\left[-\infty, \infty\right]^{31}$ & $\left[-\sqrt{2}, \sqrt{2}\right]^{242}$\\
      Total timesteps   & 5.400.000 & 5.400.000\\
      Episode length & 500 & 500\\
      
      MADDPG nr. params & 891.915 & 4.135.515 \\
      MATD3  nr. params & 1.380.918 & 7.180.250 \\
      Baseline nr. params & 560 & 3904 \\
      Hebbian nr. params & 2240 & 15616 \\
      \bottomrule
    \end{tabular}
\end{table}

\subsubsection{MARL algorithms} 
\textcolor{black}{\textbf{MADDPG} \cite{lowe2017multi}, is a version of the Deep Deterministic Policy Gradient (DDPG) algorithm that allows for learning in a swarm setting. In MADDPG, each agent maintains its own actor and critic networks. MARL methods often address the issue of non-stationarity, i.e., learned interactions between agents are inconsistent as the agents themselves adapt. This dynamic causes individual learning (where each agent learns independent of each other) to be unstable. To stabilise the critics' performance, training aggregates information with access to the global state and the actions of all agents. Agents thus have a centralised form of learning while individual actor policies only receive local information for decentralised control.}

\textcolor{black}{\textbf{MATD3} \cite{ackermann2019reducing}, this implementation extends centralised learning with decentralised control MADDPG by integrating enhancements from the single-agent TD3 algorithm. Mainly twin critic networks, target policy smoothing, and delayed actor updates. while maintaining the centralised learning decentralised control architecture. Each agent is equipped with two critics that evaluate the joint action-state space to mitigate overestimation bias, and actors are updated less frequently to stabilise learning, but subsequently result in a bigger optimisation parameter space. This approach closely mirrors the formulation in Ackermann et al. (2019) and is implemented within AgileRL for benchmarking multi-agent continuous-action tasks. }

\subsection{Swarm robotics experiment design}\label{sec:met_swarm_exp}
\subsubsection{Robot design}\label{sec:met_rob}
Our robot swarm consists of differential drive Thymio IIs without any communication capabilities, i.e., decentralised control \cite{van2022environment, van2024emergence}. The Thymio is equipped with a Raspberry Pi 5, battery and a custom sensing board for local sensing (total mass $254 \text{ gr}$); and has a maximum speed of $\pm \unit[14]{cm/s}$. In total, there are nine sensors: one light sensor, and a relative distance and heading sensor in four $90^{\circ}$ quadrants (at the \textit{front}-, \textit{left}-, \textit{back}-, \textit{right-} side of the robot). The relative distance and heading sensors only receive information of the nearest neighbour up to a distance of 2 metre (see the single Thymio in the simulator in \autoref{fig:setup}). The absence of a neighbour in a quadrant defaults the sensor to 2.01 metre distance with 0 relative heading. This information is sampled at 20Hz without memory of the previous state.

\subsubsection{Controller design}\label{sec:met_con}
We rescale the sensor input before feeding it into the NN controller ($\mathbf{s}_{in} \in [-1,1]^9$): local light intensity ($light \in [0,255]$), plus relative distance ($d_q \in [0,2.01]$), and heading ($\theta_q \in [-\pi,\pi)$) of the nearest neighbour in four quadrants). The network consists of two hidden Tanh layers of the same size as the input layer (${h}_{1}, {h}_{2} \in \mathbb{R}^9$), and a final output layer with two tanh$^{-1}$ neurons, $NN \in [-1,1]^2$. At the start, all weights are randomly initialised, with only the weights being adapted ($N_{weights}=180$, two hidden layers: $162$; and output layer: $18$), resulting in the following formulation:

\begin{align*}
    NN = \mathrm{tanh}^{-1}\left(\mathbf{W}_{out}\mathrm{tanh}^{-1}\left(\mathbf{W}_{h2}\mathrm{tanh}^{-1}\left(\mathbf{W}_{h1}\mathbf{s}_{in}\right)\right)\right) \\
    \text{$\textrm{with,} \quad  \mathbf{W}_{h1, h2} \in \mathbb{R}^{9\times9} \quad \textrm{and} \quad \mathbf{W}_{out} \in \mathbb{R}^{2\times9}$ }
\end{align*}

\begin{figure}[hpt!]
  \includegraphics[width=0.5\linewidth]{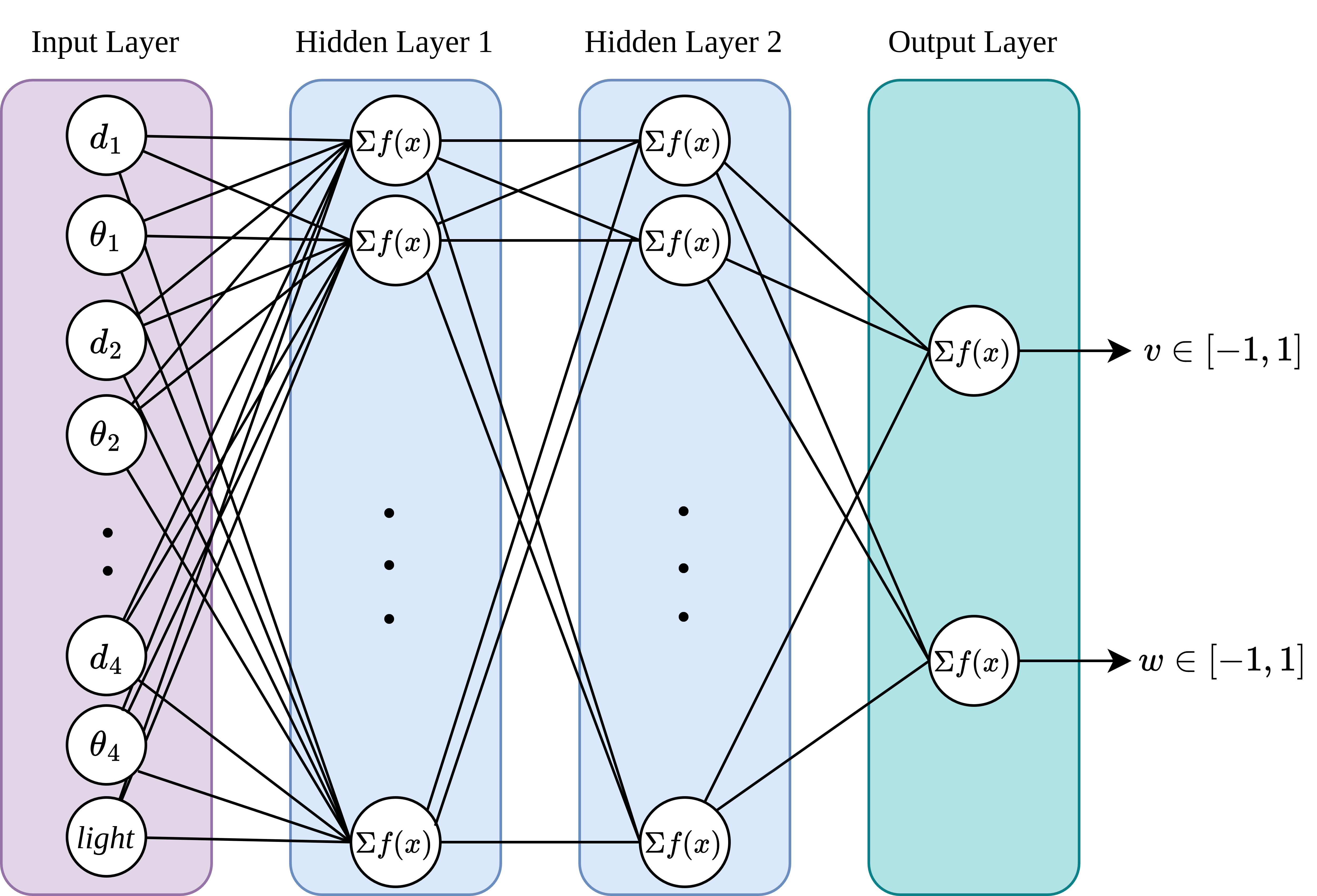}
  \centering
  \caption{\small Reservoir Neuron Network controller design.}
  \label{fig:NN}
\end{figure}

We would like to emphasize the limited capabilities of our robots: 1) there is no direct communication between agents; 2) The NN controller is memoryless, only current local sensor information is known. All in all, this grounds our notion of `limited sensing', as a single robot is incapable of estimating the gradient of the light.

\subsubsection{Source localisation task}
\begin{figure}[ht]
% \vspace{-1em}
  \centering
\begin{minipage}[c]{0.95\linewidth}
\centering
        \begin{minipage}[t]{.45\linewidth}
            \centering
            \subfloat[Isaac gym]{\includegraphics[width=\textwidth]{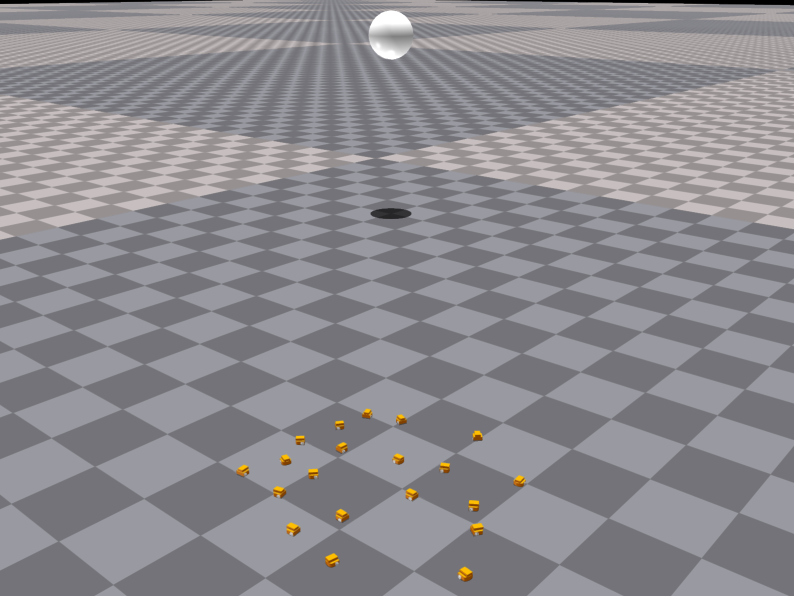}}
    \end{minipage}
\hspace{2em}
        \begin{minipage}[t]{.4\linewidth}%## ROW 10x10
            \centering
            \subfloat[Scalar field map]{\includegraphics[width=\textwidth]{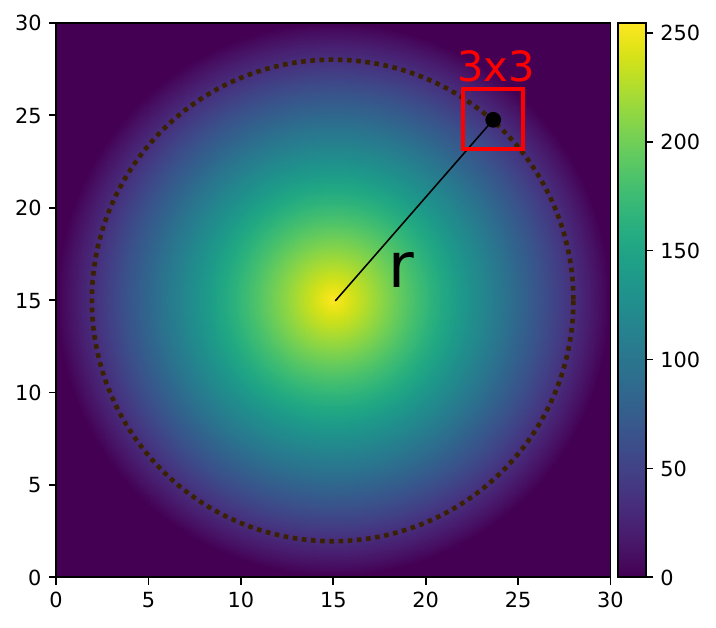}}
        \end{minipage}
\end{minipage}\vspace{-0.5em}
\caption{\small The task environment}
\label{fig:swarm_env}
% \vspace{-1em}
\end{figure}

We task our swarm to find and aggregate at the location of the highest light intensity (taken from \cite{karaguzel2020collective}), defined at the centre of an arena for a duration of 10 minutes, see \autoref{fig:setup}. For our simulation experiments we use Isaac gym where we simulate the local light intensities as a scalar field map with an arena size of $30\times30$ metre (values range between 0 and 255; see \autoref{fig:swarm_env}a). We randomly place the swarm on a circle at a fixed distance (r=12 meter) away from the centre (see \autoref{fig:swarm_env}b). At this swarm location, we randomise the position of each swarm member in a $3\times3$ metre bounding box (shown in red). We perturb the sensory readings with a Gaussian noise during simulation, based on real-world measurements (Gaussian noise model with $\mu=0$ and standard deviation: $light$=0.05, $\theta$=0.043, $d$=0.0046). In addition we simulate 20\% probability for faulty `no neighbour' sensor readings.

\subsubsection{Fitness function}
We define the fitness of a swarm as generally as possible. Therefore, we only consider information available to the agents themselves, by aggregating the average light intensity values of all members over time (see \autoref{eq:fitness}). 

\begin{minipage}[t]{.35\textwidth}\centering
\begin{equation}
{f_{trial}} = \frac{\sum_{t=0}^{T}{l_t}}{G_{max}\cdot{T}}
\label{eq:fitness}
\end{equation}
\end{minipage}% <---------------- Note the use of "%"
\begin{minipage}[t]{.5\textwidth}\centering
\begin{equation}
{l}_t = \frac {\sum_{n=1}^{20}{G_n}}{N}
\label{eq:fitness_t}
\end{equation}
\end{minipage}

% \begin{equation}
% {f_{trial}} = \frac{\sum_{t=0}^{T}{l_t}}{G_{max}\cdot{T}}\quad\text{and}\quad{l}_t = \frac {\sum_{n=1}^{20}{G_n}}{N}
% \label{eq:fitness}
% \end{equation}

Here $G_n$ is the scalar value of the grid cell in which agent $n$ is located at a time $t$. Thus the fitness at a specific time ($l_t$) is calculated as the mean scalar light value of all swarm members. Trial fitness ($f_{trial}$) is calculated by averaging all $l_t$ over total simulation time $T$. At last, we normalise using the maximum scalar value $G_{max}$, always equal to $255$ for all experiments. A theoretical maximum fitness of 1 can only be achieved if all swarm members would be stacked at the centre, for the full duration of the run. We repeat a single trial three times from random equidistant starting positions (away from the centre) and assign the final fitness $F$ as the median performance $F=\texttt{MEDIAN} (f_1, f_2, f_3)$.

\subsection{Evolutionary Algorithm}\label{sec:met_ese}
We optimise our swarm using Covariance Matrix Adaptation Evolution Strategy (CMA-ES, \cite{hansen2001cmaes}). We evolve our controller for 100 generations with 30 individuals (note that each individual defines a \textit{single swarm} here). A general overview of CMA-ES is provided below. We run 10 repetitions of all controller types, and use the overall best controller for further analysis. The conditions of the experiments are presented in \autoref{tab:parameters_Special_s}. CMA-ES is used for evolving the Hebbian learning rules in the Hebbian condition, and directly evolving the weights in the Baseline/Baseline-A condition.

% \begin{algorithm}[h]\small
% \SetAlgoNlRelativeSize{0}
% \caption{\small CMA-ES Pseudocode\label{alg:cmaes}}
% \KwData{Initial mean $\mathbf{m}_0 = 0$, initial covariance $\sigma_0$, population size $\lambda = 30$ solutions $\{\mathbf{x}_1, \mathbf{x}_2, \ldots, \mathbf{x}_\lambda\}$ with genotype $\mathbf{x}_i \in \mathbb{R}^{dims}$}
% \KwResult{Optimal solution $\mathbf{x}^*$}
% % $gen \leftarrow 0$

% $N_{gen}\leftarrow 100$

% $N_{repeats}\leftarrow 3$

% \For{$gen$ in $N_{gen}$}{
%   \For{$\mathbf{x}_i$ in  $\lambda$}{    
%     \For{$rep$ in $N_{repeats}$}{
%         $f_{rep} \leftarrow \textrm{SIMULATE}(\mathbf{x}_i)$ \hfill {\textit{10min simulation}}
%     }

%     $F_i\leftarrow \textrm{MEDIAN}(f_1, \dots, f_{N_{repeats}})$ \hfill \textit{Fitness}
%   }
%   Sort solutions by fitness
  
%   Select the top-performing solutions to form a weighted mean
  
%   $\mathbf{m}_{t+1} \leftarrow \sum_{i=1}^{w} w_i \mathbf{x}_i$, where $w_i$ are weights
  
%   Update covariance matrix $\mathbf{C}_{gen+1}$
  
%   Generate a new population covariance matrix $\mathbf{C}_{\text{new}}$ based on the selected solutions
  
%   $\mathbf{C}_{gen+1} \leftarrow \mathbf{C}_{\text{new}}$
  
%   Sample $\lambda$ solutions $\{\mathbf{x}_1, \mathbf{x}_2, \ldots, \mathbf{x}_\lambda\}$ using $\mathbf{C}_{gen+1}$
% }
% $\mathbf{x}^* \leftarrow$ Best solution found
% \Return{$\mathbf{x}^*$}
% \end{algorithm}

\begin{table}[htp!]
\small
\caption{\small Evolving swarm experiment parameters}
\centering
\begin{tabular}[b]{{p{0.15\linewidth} p{0.25\linewidth} p{0.53\linewidth}}}
                 & Value           & Description\\
\toprule
Runs      &  10             & Number of runs of our experiment \\ 
% \bottomrule 
% \toprule
\midrule
\multicolumn{3}{l}{\textbf{Learning task}: \textit{collective gradient sensing}} \\
\midrule
Swarm size   & 20    & Number of robots in a swarm \\
% Ratio        & 1:1   & sub-group division          \\ 
r        & 12   & Spawn distance from centre (meters)          \\ 
Arena type  & centre     & Environment: centre	    \\
Eval. time   & 10    & Test duration in minutes    \\ 
% \bottomrule 
% \toprule
\midrule
\multicolumn{3}{l}{\textbf{Optimiser}: \textit{CMA-ES}}     \\
\midrule
Genotype      & $U[-1,1]$ & Initial sampling $\mathbf{x}_{init}$   \\
$N_{ABCD}$ & $720, \in [-5,5]$ & Number of rules (evolved during Hebbian) \\
$N_{weights}$ & $180, \in [-\infty, \infty]$ & Number of weights (evolved during Baseline) \\
$\lambda$   & 30        & Population size          \\
$N_{gen}$ & 100       & Termination condition            \\ 
$\sigma_0$  & 1.0       & Initial covariance value              \\ 
$N_{repeats}$     & 3         & Number of repeats per individual \\
\bottomrule 
\end{tabular}
\label{tab:parameters_Special_s}
\end{table}

\subsection{Hebbian Learning}\label{sec:met_heb}
Hebbian learning locally updates the weights inside of the NN, according to correlations between the activations of the associated neurons. In our case we follow the correlation ABCD-rule update method formulated in \autoref{eq:abcd}. With neuron $i$ and $j$ are connected through weight $w_{i,j}$, and the respective neuron activations $n_{i}$, $n_{j}$ are inducing an update $\delta w_{i,j}$; scalars $A_{i,j},B_{i,j},C_{i,j},D_{i,j} \in [-5,5]$ are the correlation rules subject to evolution. The learning rate $\mu$ is a constant which is the same for every weight ($0.1$). These update rules are applied update the weights after the controller inference step. 

During a trial, each swarm member receives the same ABCD rules and initialises the NN controller randomly $U[-1,1]$. Weight adaptation takes place locally per robot (with the same set of learning rules). Note that, although the learning rules a generalised, the updates are based on local inputs, thus resulting in unique weight dynamics for each individual robot. 
For Hebbian learning, we optimise the ABCD-rules for each weight resulting four times as many parameters as the number of weights ($N_{params} = 4\cdot|\mathcal{W}|$. We initialise our NN weights randomly using uniform sampling between -1 and 1 ($U[-1,1]$), our ABCD-rules $U[-0.1, 0.1]$, and the CMA-ES algorithm with $\sigma_0=1$. 

\textcolor{black}{
\subsection{Baseline(-A) controller}\label{sec:met_baseline}
The \textit{Baseline controller} is `classic' approach in swarm robotics where each member of the swarm gets assigned the same NN controller. Instead of learning rules that update the weights during the trials, we now update the weights directly between the trials for the Baseline. The same CMA-ES is used to now optimise the weight space (see \autoref{tab:parameters_Special_s}). We decided to maintain the same network architecture as the Hebbian method to ensure that the Baseline can express the same functionalities. \autoref{APP_NAD} shows that other neural designs perform similarly well but increasing the weights significantly reduces computation speed. \\ 
The \textit{Baseline-A} controller is adapted from \cite{van2024emergence}. This method requires us to define the number of sub-groups beforehand (two were chosen) that co-evolve different NNs in the same environment using CMA-ES. At the start of a trial we assign each member to one of the sub-groups randomly (with a 50-50 split). Both groups are unaware of the sub-group division and perceive the same sensory input. In the end, this allows for specialisation of the two different NNs, while inducing collective emergent behaviour through the limited capability of the individuals. The fitness of the two network is the combined mean average light intensity (see \autoref{eq:fitness}) of the whole swarm. Similar to the Hebbian and Baseline experiment we run 10 runs in total and pick the best overall controller.}

The two neural networks serve as a behavioural repertoire that is switched during deployment through a probabilistic finite state machine. This could be beneficial if each subgroup behviour has specialised for different parts of the task (e.g. collective gradient sensing vs. aggregation at the centre). The probabilistic state machine is tuned to bias the NN state towards a swarm distribution that performs best at a certain local light sensor value. The optimal state probabilities are obtained by retesting the controller for different swarm ratios, at different distances away from the light source, \autoref{tab:ratio_robustness}. 

\newcommand*{\MaxNumber}{0.21}
\newcommand*{\MidNumber}{0.639}
\newcommand*{\MinNumber}{0.83}
\newcommand{\mytextcolor}[2]{\ifdim#1pt<\MidNumber pt\textcolor{gray!10}{#1}\else #1\fi}
\newcommand{\circletextcolor}[2]{\ifdim#1pt<\MidNumber pt\textcolor{gray!10}{\circlered{#2}{#1}}\else \circlered{#2}{#1}\fi}

\newcommand{\ApplyGradient}[2]{
    \FPeval{\result}{100*(#1-\MinNumber)/(\MaxNumber-\MinNumber)}
    \centering
    \edef\x{\noexpand\cellcolor{black!\result}}
    \x\mytextcolor{#1}
}
\newcommand{\ApplyGradientCircle}[2]{
    \FPeval{\result}{100*(#1-\MinNumber)/(\MaxNumber-\MinNumber)}
    \centering
    \edef\x{\noexpand\cellcolor{black!\result}}
    \x\circletextcolor{#1}{#2}
}
\newcommand{\ApplyGradientCircleA}[2]{\ApplyGradientCircle{#1}{a}}
\newcommand{\ApplyGradientCircleB}[2]{\ApplyGradientCircle{#1}{b}}
\newcommand{\ApplyGradientCircleC}[2]{\ApplyGradientCircle{#1}{c}}
\newcommand{\ApplyGradientCircleD}[2]{\ApplyGradientCircle{#1}{d}}
\newcommand{\ApplyGradientCircleE}[2]{\ApplyGradientCircle{#1}{e}}
\newcommand{\ApplyGradientCircleF}[2]{\ApplyGradientCircle{#1}{f}}
\newcolumntype{A}{>{\collectcell\ApplyGradientCircleA}p{0.5cm}<{\endcollectcell}}
\newcolumntype{B}{>{\collectcell\ApplyGradientCircleB}p{0.5cm}<{\endcollectcell}}
\newcolumntype{C}{>{\collectcell\ApplyGradientCircleC}p{0.5cm}<{\endcollectcell}}
\newcolumntype{D}{>{\collectcell\ApplyGradientCircleD}p{0.5cm}<{\endcollectcell}}
\newcolumntype{E}{>{\collectcell\ApplyGradientCircleE}p{0.5cm}<{\endcollectcell}}
\newcolumntype{F}{>{\collectcell\ApplyGradientCircleF}p{0.5cm}<{\endcollectcell}}
\newcolumntype{R}{>{\collectcell\ApplyGradient}p{0.5cm}<{\endcollectcell}}
\newcolumntype{Q}{>{\collectcell\ApplyGradient}p{0.4cm}<{\endcollectcell}}
\begin{table}[h]
  \centering
  \renewcommand{\arraystretch}{1.8}
  % \begin{tabular}{ *{8}{Q} }
  %   0.1 & 0.2 &
  %   0.3 & 0.4 & 0.5 &
  %   0.6 & 0.7 & 0.8 \\
  %   \multicolumn{1}{c}{\vspace{-.7em}} \\
  % \end{tabular}
  \begin{tabular}{
    c  *{5}{R}
  }
  \toprule
      \multicolumn{1}{c}{ratio~$\rightarrow$}& 
      \multicolumn{1}{c}{\cellcolor{green!30}$4:0$} &
      \multicolumn{1}{c}{$3:1$} & 
      \multicolumn{1}{c}{$2:2$} & 
      \multicolumn{1}{c}{$1:3$} &
      \multicolumn{1}{c}{\cellcolor{red!30}$0:4$} \\
      % & $p<0.05$ \\
    \midrule
    
    $r_{dist}=0.00$ & 0.69 & 0.73 & 0.75 & 0.80 & \multicolumn{1}{A}{0.82} \\% & \cellcolor{green!30}*** \\ % p=2.368e-23
    $r_{dist}=0.25$ & 0.67 & 0.71 & 0.73 & \multicolumn{1}{B}{0.78} & 0.78\\% & \cellcolor{green!30}*** \\ % p=2.368e-23
    $r_{dist}=0.50$ & 0.64 & 0.70 & \multicolumn{1}{C}{0.71}  & 0.70 & 0.61 \\% & \cellcolor{green!30}*** \\ % p=1.956e-08
    $r_{dist}=0.75$ & 0.60 & \multicolumn{1}{D}{0.64} & 0.59  & 0.52 & 0.38 \\% & \cellcolor{green!30}*   \\ % p=0.0108
    $r_{dist}=1.00$ & 0.47 & \multicolumn{1}{E}{0.51} & 0.41 & 0.26 & 0.21 \\% & \cellcolor{red!30}*   \\ % p=0.0116
    \bottomrule            
  \end{tabular}
  \begin{tikzpicture}[remember picture,overlay]
    % \draw[red,thick] (a) circle[x radius=4mm,y radius=2mm];
    \draw[red,thick] (a) ++(-4.6mm,-2.5mm) rectangle ++(8.9mm,5.3mm);
    \draw[red!70, dashed] (a) ++ (-14.8mm,-2.5mm) rectangle ++(8.9mm,5.3mm);
    \draw[red,thick] (b) ++(-4.6mm,-2.5mm) rectangle ++(8.9mm,5.3mm);
    \draw[red!70, dashed] (b) ++(5.5mm,-2.5mm) rectangle ++(8.9mm,5.3mm);
    \draw[red!70, dashed] (c) ++ (-14.8mm,-2.5mm) rectangle ++(8.9mm,5.3mm);
    \draw[red,thick] (c) ++(-4.6mm,-2.5mm) rectangle ++(8.9mm,5.3mm);
    \draw[red!70, dashed] (c) ++(5.5mm,-2.5mm) rectangle ++(8.9mm,5.3mm);
    \draw[red,thick] (d) ++(-4.6mm,-2.5mm) rectangle ++(8.9mm,5.3mm);
    \draw[red,thick] (e) ++(-4.6mm,-2.5mm) rectangle ++(8.9mm,5.3mm);
  \end{tikzpicture}
  \hspace{1mm}
  \caption{\small Average fitness values ($N=60$) of retesting the best swarm with different sub-group ratios (green:red) from different starting distances ($r_{dist}$ = distance to the light source). Sub-group ratios vary from solely sub-group 1 (red) to solely sub-group 2 (green). %Deviation from the mean varies from $0.01$ to $0.14$, depending on the configuration. 
  Solid red boxes indicate best ratio at a given $r_{dist}$, while the dashed boxed indicate no statistically significant differences with respect to the maximum.
    \label{tab:ratio_robustness}
  }
  % \vspace{-1em}
\end{table}

\textnormal
The best-performing ratios define how likely it is for a robot to select one of the two NN for their next control sequence. Given the best swarm ratios at a given light intensity (corresponding to the distance r from the light) we assign a probability to switch NN such that (as a whole) the swarm adapts its behaviour optimally. In this way, the \textit{Baseline-A} adaptation mechanism only requires local information, thus maintaining the distributed control ability. Thresholds are assigned on the light intensities at $r_{dist} = \{0.125, 0.375, 0.625, 0.875\}$. This results in the following function for the probability of expressing the first green sub-group behaviour ($P_{green}$). 

{\small
\[   
P_{green}(light) = 
     \begin{cases}
       \text{1.0} &\quad\text{if $light$} >229\\
       \text{0.75} &\quad\text{if $light$} \in (178,229] \\
       \text{0.50} &\quad\text{if $light$} \in (127,178]\\
       \text{0.25} &\quad\text{if $light$} \le 127\\
     \end{cases}
\]
}
Here, $light$ refers to the current light intensity measured by the robot ([0, 255]) and $P_{black} = 1 - P_{green}$.

\subsection{Hebbian weight dynamics experiment design}
We perturb the light source during deployment to elucidate the dynamics of the weight during deployment (after 300 seconds). A comparison is made with a `static' environment which is exactly the same as the training environment. In the dynamic environment we perturb the light source to the (3,3) metre position (see \autoref{fig:dyn_v_stat}a right side). During each trial, we measure:

\subsubsection{Average light intensity}
We track the average sensed light intensity of the swarm over time, see \autoref{eq:fitness_t}. Note that this metric is not the same as the fitness used during evolution (which would be the time average of the accumulated values). Instead this metric provides insight on how the swarm readapts during deployment.

\subsubsection{Average mean weight autocorrelation}
For each individual weight we calculate the autocorrelation over the whole run using \texttt{numpy.correlate}, $c(\tau)$ with $\tau$ indicating the time-lag. We average the correlations over all NN weights ($N_{weights}=180$) for all agents ($N_{agents}=20$) at each time-step to derive the average mean weight autocorrelation (see \autoref{eq:autocor}). Results are shown for positive time lags up to $\tau=N_{timesteps}-1$, to measure the similarity of the weight time-series at $t_0:t_{-\tau}$ with the weight time-series at $t_\tau:t_{timesteps}$. 

\small{
\begin{equation} \label{eq:autocor}
    \bar{c}(\tau) = \dfrac{1}{N_{agents}\cdot N_{weights}}\sum_{n_{a}}^{N_{agents}} \sum_{n_{w}}^{N_{weights}} c(\tau, w_{n_{w}}^{n_{a}}) \quad with \quad c(\tau, w) = \sum_t w(t+\tau) \cdot w(t)
\end{equation}
}

\subsubsection{Analysis of swarm heterogeneity}
We analyse the swarm heterogeneity by measuring the average weight variance between the NN weights of each agent using \texttt{numpy.std}. We calculate the standard deviation over the members of the swarm (20 NNs) or each individual weight, and average over all weights (180 weights), see \autoref{eq:meanstd}. \textcolor{black}{If each robot's network is the same --Note that the weights inside the network can differ-- than the mean STD equals 0.}

\small{
\begin{equation} \label{eq:meanstd}
    \overline{\textrm{std}}(t) = \dfrac{1}{ N_{weights}} \sum_{n_{w}}^{N_{weights}} \textrm{std}(\mathbf{w}_{n_w}(t)) \quad with \quad \mathbf{w}_{n_w}(t) = [w^{n_a}_{n_w}(t), \dots, w^{N_{agents}}_{n_w}(t)]
\end{equation}
}

\subsection{Real world swarm setup}\label{sec:met_real} 
Finally, we transfer the best version (out of 10 simulated evolutionary runs) of the Baseline; Baseline-A; and Hebbian controller, in a real robot swarm. 
We extend the Thymios' capabilities with custom on-board sensing for decentralised control (see \autoref{fig:setup} bottom right). For localisation, we utilise the Crazyflie Lighthouse V2 Positioning System \cite{taffanel2021lighthouse} (also see \url{https://www.bitcraze.io/documentation/system/positioning/ligthouse-positioning-system}, accessed 6-2-2025), which implements an optical beacon that shines a laser that is captured by 4 photo-diode receivers on our board to provide full body pose. Light sensing is achieved through a TEMT6000 \url{https://www.vishay.com/docs/81579/temt6000.pdf} light sensor circuit, which changes the output voltage depending on the light intensity in the environment. Each board communicates positions and headings via a radio communication channel using a P2P API(\url{https://www.bitcraze.io/documentation/repository/crazyfliefirmware/master/functional-areas/p2p_api/}, accessed 6-2-2025). \textcolor{black}{The heading and distance information is used to calculate the nearest neighbour per quadrant on the CrazyFlie system, which feeds this information to a Raspberry Pi 5 \url{https://www.raspberrypi.com/documentation/computers/raspberry-pi.html} (accessed 6-2-2025). The GPS is used solely to provide local information onboard each robot, maintaining decentralization.} The Pi, responsible for control, only receives local neighbour data to sample the (Hebbian) NN controller, and sends the motor commands to the Thymio. This hardware setup results in a complete onboard system for distributed control. Code base for the hardware implementation is available at: \url{https://github.com/fudavd/CrazyThymio}. 

Each controller is tested in a $8\times8$ metre environment with a light source at the centre producing a gradient. We start the Thymios at the edge of the light ($\sim$3.5 meter) and log the sensed light value, position, and yaw angle using a serial radio communication channel. At the start of a run, a central machine sends a start command that activates the wheels randomly for 5 seconds to position the swarm in a random position and orientation. Subsequently, we start a run for 10 minutes in which each swarm member is controlled locally. We test each controller 3 times to get a robust estimate of its behaviour. In the end, we compare the fitness over time and the collective behaviour of the swarm.

\section{Data Availability}
The raw and processed data generated in this study will be deposited in a database \url{https://dataverse.nl/}. 

\section{Code Availability}
Evolutionary, and simulated experiments code: \url{https://github.com/fudavd/EC_swarm/tree/Hebbian}.

\section*{Acknowledgments}
The work of EF has been supported by the NYUAD Center for Artificial Intelligence and Robotics, funded by Tamkeen under the NYUAD Research Institute Award CG010.

\section*{Author Contributions Statement}
\addcontentsline{toc}{section}{Author Contributions Statement}
Fuda van Diggelen: main contributing author, developed code base, conducted all experiments.\newline
Tugay Alperen Karag{\"u}zel: developed hardware with integrated software, conducted real-world experiments.\newline
Andrés García Rincón: developed hardware with integrated software, conducted real-world experiments.\newline
Agoston Eiben: contributed to writing, conducted real-world experiments.\newline
Dario Floreano: contributed to writing, experimental design.\newline
Eliseo Ferrante: contributed to writing, experimental design.

\section*{Competing Interests Statement}
\addcontentsline{toc}{section}{Competing Interests Statement}
The authors declare no competing interests.

%%===========================================================================================%%
%% If you are submitting to one of the Nature Portfolio journals, using the eJP submission   %%
%% system, please include the references within the manuscript file itself. You may do this  %%
%% by copying the reference list from your .bbl file, paste it into the main manuscript .tex %%
%% file, and delete the associated \verb+\bibliography+ commands.                            %%
%%===========================================================================================%%

\bibliography{sn-bibliography}% common bib file
%% if required, the content of .bbl file can be included here once bbl is generated
%%\input sn-article.bbl

%%%%%%%%%%%%%%%%%%%%%%%%%%%%%%%%%%%%%%%%%%%%%%%%%%%%%%%%%%%%%%%%%%%%%%%%

%%% The next two lines define, first, the bibliography style to be 
%%% applied, and, second, the bibliography file to be used.

\newpage
\begin{appendices}

\section{Appendix}\label{secA1}
\subsection{Additional results}\label{APP_add_res}
After our evolutionary experiments we obtain the overall best controller over 10 runs. In swarm robotics robustnesses during deployment (after training) is often considered in terms of \textit{Scalability} (performance when swarm size changes), and \textit{Flexibility} (performance when environments change). In our Scalability experiments (Appendix \ref{APP:scalability}), we place our swarm in the same light source environment but with different swarm sizes. Here, performance comparison is made within the same controllers between the original swarm size of 20 Thymios. The experimental parameters we used in all the validation experiments are described in \autoref{tab:parameters_Special_s_val}. In the Flexibility environment (Appendix \ref{APP:flexibility}), we place our swarm of size 20 Thymios in different gradient maps; and compare relative performance between controllers within the same map.

\renewcommand{\arraystretch}{0.6} % Default value: 1
\begin{table}[ht]
\small
\caption{\small Validation experiment parameters}
\centering
\begin{tabular}[b]{{p{0.18\linewidth} p{0.20\linewidth} p{0.43\linewidth}}}
 & Value & Description\\
\toprule     
Repetitions      &  60    & Number of runs per experiment \\ 
Statistics       & \textit{t}-test & Statistical test \\ \bottomrule 
\toprule
\multicolumn{3}{l}{\textbf{Scalability}: \textit{Different swarm size}} \\ \midrule
Swarm size 			 & 10/20/50/100     & Number of robots in a swarm 	\\
Arena type 			 & centre    & Environment 	\\
Eval. time  & 10    & Test duration in minutes \\ 
\bottomrule 
\toprule
\multicolumn{3}{l}{\textbf{Flexibility}: \textit{Different environments}} \\ \midrule
Swarm size 			 & 20     & Number of robots in a swarm 	\\
% Ratio      &2:2/adaptive & sub-group division \\
Arena type 			 & lin./bi\nobreakdash-mod./ros.     & Environment 	\\
Eval. time  & 10    & Test duration in minutes \\ 
\bottomrule 
\end{tabular}
\label{tab:parameters_Special_s_val}
\end{table}

\subsection{Scalability of best controller}\label{APP:scalability}
We consider the best controllers from \autoref{fig:Results} (and additional NN architectures) to investigate the flexibility of our approach for different swarm sizes (see \autoref{tab:scal}). These  controllers are tested in different swarms with varying sizes $\left[10, 50, 100\right]$ robots, and compare it with the performance of the original 20 robots in the same circular gradient environment. Each experimental condition is repeated 60 time ($N=60$). Swarm size perturbation is expected to decrease performance as the swarms are optimised for the current size.

The Baseline shows a significant drop in performance when changing the swarm size from 20 robots ($0.61\pm0.04$) to any other swarm size. The smallest Baseline performance drop is with respect to swarm size 50 ($0.56 \pm 0.018$), which is significantly lower $p\le0.001$. This is not the case for Hebbian learning which significantly outperforms the Baseline in all conditions ($p\le0.001$), and has a smaller performance drop overall (both absolute and relative) with respect to the swarm size of 20 robots ($0.77\pm0.026$). Interestingly unlike the Baseline method, for the Hebbian case, we find that changing the swarm size maintains performance more effectively. On an aggregate, Hebbian learning outperforms the Baseline significantly on the scalability experiments with Bonferroni correction ($p\le0.01/\alpha$ where $\alpha=3$, $df=358$).

\begin{table}[ht]\small
    \centering
    \begin{tabular}{l cccc}
       \toprule
       \textbf{Scalability} & \multicolumn{3}{c}{\textit{Perturbed swarm size}} & original\\
       N=60  &  10 & 50 & 100 & \textit{20}\\
         \midrule
       Baseline  &  $0.42\pm0.013$ & $0.56 \pm 0.018$ & $0.47 \pm 0.016$ & $0.61 \pm 0.035$\\
       % Adaptive \cite{van2024emergence}  &  $0.40\pm0.012$ & $0.64 \pm 0.016$ & $0.58 \pm 0.017$\\ $0.68\pm0.04$
       
       % H-single  &  $0.24\pm0.039$ & $0.39 \pm 0.018$ & $0.57 \pm 0.018$\\
       Baseline-A  &  $0.40\pm0.012$ & $0.64 \pm 0.016$ & $0.58 \pm 0.017$ & $0.68 \pm 0.036$\\
       
       Hebbian-1  &  $0.24\pm0.039$ & $0.39 \pm 0.018$ & $0.57 \pm 0.018$ & $0.22 \pm 0.031$\\
       Recurrent  &  $0.20\pm0.041$ & $0.19 \pm 0.031$ & $0.19 \pm 0.038$ & $0.61 \pm 0.029$ \\

       Hebbian  &  $\hphantom{^*}\bm{0.77\pm0.020}^*$ & $\hphantom{^*}\bm{0.72\pm0.033}^*$ & $\hphantom{^*}\bm{0.65\pm0.024}^*$ & $\hphantom{^*}\bm{0.77\pm0.026}^*$\\
       \bottomrule
    \end{tabular}
    \caption{\small Scalability experiments in the circular gradient environment (results shown are averaged over 60 repetitions, $N=60$ with $\pm\text{STD}$).}
    \label{tab:scal}
\end{table}

\subsection{Flexibility of best controller}\label{APP:flexibility}
\begin{figure}[ht]
  \centering
\begin{minipage}[c]{0.85\linewidth}
\centering
        \begin{minipage}[t]{.3\linewidth}
            \centering
            \subfloat[(a) Linear]{\includegraphics[trim={8em 2em 8em 1em},clip,width=\textwidth]{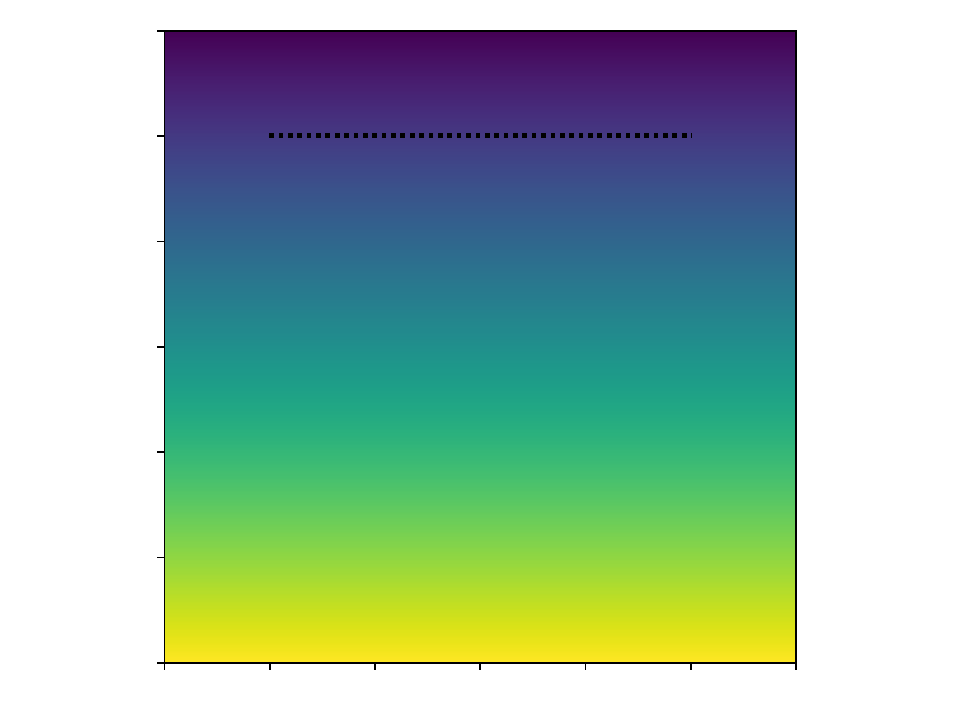}}
        \end{minipage}
        \begin{minipage}[t]{.3\linewidth}%## ROW 10x10
            \centering
            \subfloat[(b) Bi-modal]{\includegraphics[trim={8em 2em 8em 1em},clip, width=\textwidth]{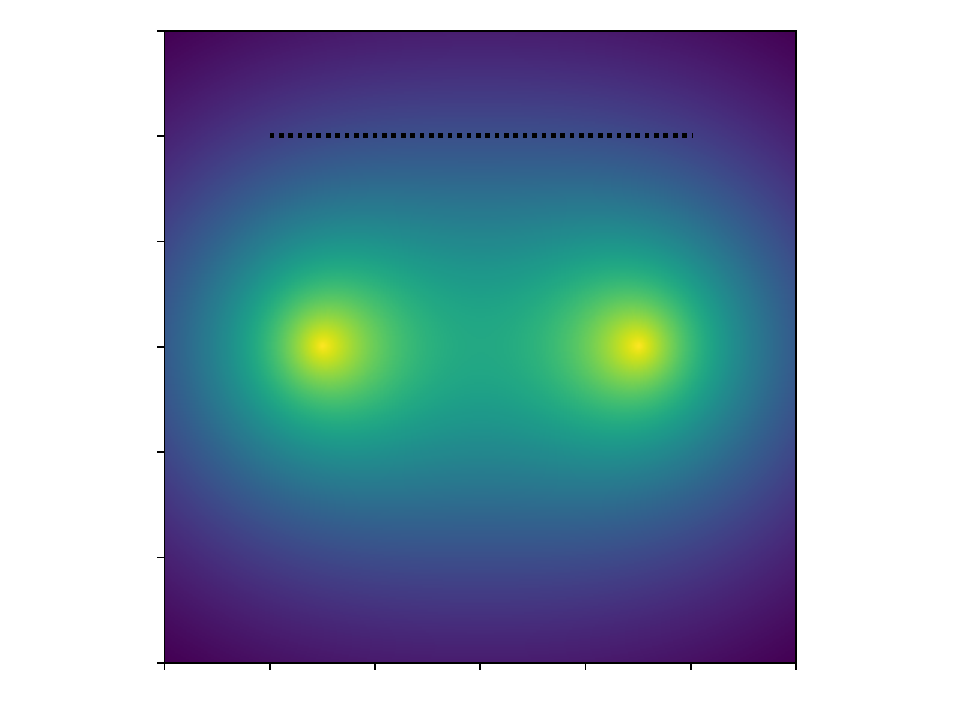}}
        \end{minipage}
        \begin{minipage}[t]{.3\linewidth}
            \centering
            \subfloat[(c) Rosenbrock]{\includegraphics[trim={8em 2em 8em 1em},clip, ,width=\textwidth]{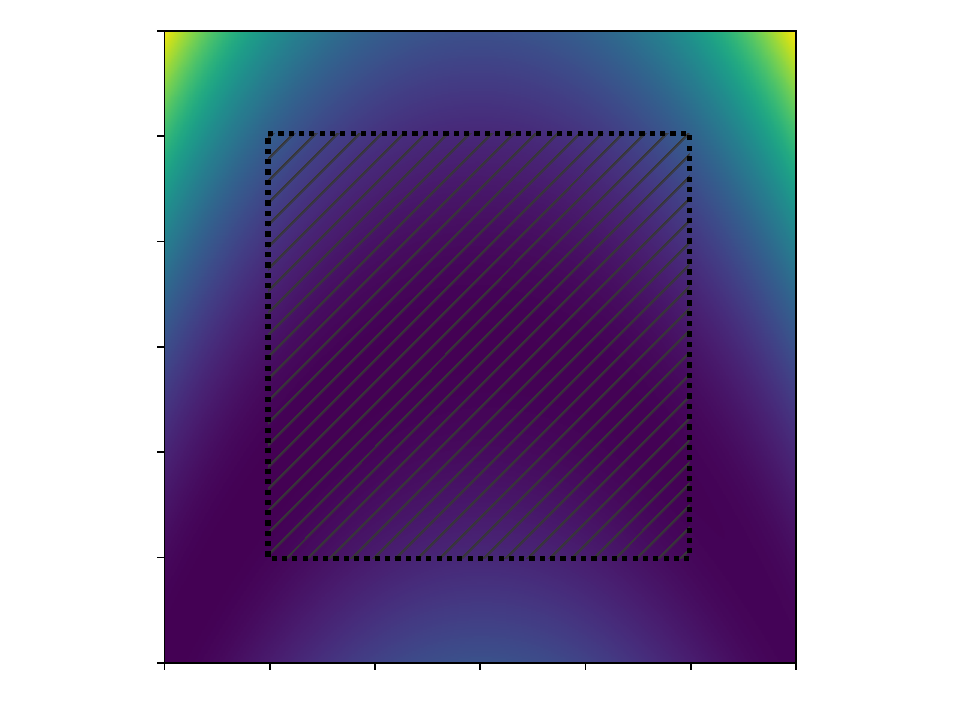}}
    \end{minipage}
\end{minipage}
\caption{\small Validation environments. Black striped line indicate the randomised initial location of the swarm. Striped box in (c) indicate the area of random initialisation.}
\label{fig:val_env}
\end{figure}

In our flexibility experiments, we initialise the swarm (of size 20) in an unseen environment, by changing the gradient map $\left[\text{Linear},\text{Bi-modal},\text{Rosenbrock}\right]$ as shown below. \textcolor{black}{Please note that comparison between arena types are unfair since each arena pertains its own maximum fitness.} Each arena presents a different challenge: \textit{Linear} poses a less salient gradient stretched out over the full arena (\autoref{fig:val_env}a); \textit{Bi-modal}, requires collective decision on where to go (\autoref{fig:val_env}b); \textit{Rosenbrock}, the Rosenbrock function is taken from a classic a non-linear minimisation problem \cite{clerc1999swarm} and has a shallow centre with a local maximum and two distant maxima (\autoref{fig:val_env}c).

We consider flexibility in the form of resistance with respect to environmental perturbations (see \autoref{tab:rob}). In this case, we employ the best controllers from \autoref{fig:Results} and test their performance in different gradient maps, $\left[\text{Linear},\text{Bi-modal},\text{Rosenbrock}\right]$ (see \autoref{fig:val_env} for the gradient map). The \textit{Linear} problem introduces less pronounced gradient, with a wide optimum. This arena might impose difficulties finding the gradient direction and stabilising behaviour at the optimum. The \textit{Bi-modal} problem introduces two optima. The best behaviour requires the swarm to find the nearest optimum and collectively aggregate at this peak. The \textit{Rosenbrock} problem contains a shallow minimum and local optima. Each experimental condition is repeated 60 time ($N=60$). 

More pronounced than the scalability experiments the Hebbian learning significantly outperforms the Baseline method during the flexibility experiments in all conditions ($p\le0.001$). Beside this consistent outperformance of Hebbian for every environment, we can also see a significantly smaller performance drop with respect to the original circular environment (both absolute and relative drop for all environments, $p\le0.001$). Overall, Hebbian learning outperforms the Baseline significantly on the flexibility experiments with Bonferroni correction ($p\le0.01/\alpha$ where $\alpha=3$, $df=358$).

\begin{table}[ht]\small
    \centering
    \begin{tabular}{l cccc}
       \toprule
       \textbf{Flexibility} & \multicolumn{3}{c}{\textit{Perturbed arena type}} & original\\
       N=60 & {Linear} &  {Bi-modal}  &  {Rosenbrock} & \textit{circular}\\
         \midrule
       Baseline  & $0.18 \pm 0.050$ &  $0.28 \pm 0.10$ & $0.15 \pm 0.18$ & $0.61 \pm 0.037$\\
       % Adaptive \cite{van2024emergence}  & $0.64 \pm 0.016&  $0.40\pm0.012$ $ & $0.58 \pm 0.017$\\ $0.68\pm0.04$
       
       % H-single & $0.39 \pm 0.018 &  $0.24\pm0.039$ $ & $0.57 \pm 0.018$\\
       Baseline-A  &  $0.35\pm0.032$ & $0.38 \pm 0.080$ & $0.14 \pm 0.032$ & $0.68 \pm 0.036$\\
       
       Hebbian-1  &  $0.35\pm0.023$ & $0.21 \pm 0.071$ & $0.13 \pm 0.130$ &  %$0.83 \pm 0.040$ 
       $\hphantom{^*}\bm{0.83\pm0.040}^*$
       \\
       Recurrent  & $0.31 \pm 0.033$ & $0.21\pm0.052$ &  $0.08 \pm 0.101$ & $0.61 \pm 0.029$ \\
    
       Hebbian  & $\hphantom{^*}\bm{0.57 \pm 0.043}^*$ &  $\hphantom{^*}\bm{0.57 \pm 0.037}^*$ & $\hphantom{^*}\bm{0.41 \pm 0.19}^*$ & $0.77\pm0.021$\\
       \bottomrule
    \end{tabular}
    \caption{\small Flexibility experiments with swarm size 20 (results shown are averaged over 60 repetitions, $N=60$ with $\pm\text{STD}$).}
    \label{tab:rob}
\end{table}

% \begin{table}[ht]\small
%     \centering
%     \begin{tabular}{l ccc}
%        \toprule
%        \textbf{Scalability} & \multicolumn{3}{c}{\textit{Swarm size}}\\
%        N=60  &  10 & 50 & 100\\
%          \midrule
%        Baseline  &  $0.42\pm0.013$ & $0.56 \pm 0.018$ & $0.47 \pm 0.016$\\
%        Adaptive \cite{van2024emergence}  &  $0.40\pm0.012$ & $0.64 \pm 0.016$ & $0.58 \pm 0.017$\\
       
%        % H-single  &  $0.24\pm0.039$ & $0.39 \pm 0.018$ & $0.57 \pm 0.018$\\
       
%        Hebbian  &  $\hphantom{^*}\bm{0.77\pm0.020}^*$ & $\hphantom{^*}\bm{0.72\pm0.033}^*$ & $\hphantom{^*}\bm{0.65\pm0.024}^*$ \\
%        \bottomrule
%        \toprule
%        \textbf{Flexibility} & \multicolumn{3}{c}{\textit{Arena type}}\\
%        N=60 &  {Bi-modal} & {Linear} &  {Rosenbrock} \\
%        \midrule
%        Baseline  &  $0.28 \pm 0.10$ & $0.18 \pm 0.050$ & $0.15 \pm 0.18$\\
%        Adaptive \cite{van2024emergence}  &  $0.38\pm0.08$ & $0.35 \pm 0.032$ & $0.14 \pm 0.19$\\
       
%        % H-single  &  $0.21\pm0.07$ & $0.35 \pm 0.023$ & $0.13 \pm 0.13$\\
%        Hebbian  &  $\hphantom{^*}\bm{0.57 \pm 0.037}^*$ & $\hphantom{^*}\bm{0.57 \pm 0.043}^*$ & $\hphantom{^*}\bm{0.41 \pm 0.19}^*$ \\
%        \bottomrule
       
%     \end{tabular}
%     \caption{\small Validation experiments ($N=60$)}
%     \label{tab:res_val}
% \end{table}

Combining all the results from \autoref{tab:scal}\&\ref{tab:rob}, on an aggregate Hebbian learning significantly outperforms the Baseline ($N=360$). The aggregated mean performance is statistically significantly different with Bonferroni correction ($p\le0.001/\alpha$ where $\alpha=6$, $df=718$). Additional ablation test were made with different control architectures, presented in Appendix \ref{APP:comp_plus}. For all methods, Hebbian learning significantly outperformed in overall performance, Scalability and Flexibility (all comparisons $p\leq0.001$).

\subsection{Additional controller comparison} \label{APP:comp_plus}
We conducted additional experiments to further investigate the following properties 1) Collaboration ; 2) Memory/state propagation. This results in evolving two additional control strategies: 1) a Hebbian controller evolved with a single agent in the arena; 2) A recurrent NN where the output of the previous time step is presented as additional input, (resulting in an input layer of size 9+2, Hidden layer of size 9-, and output layer of size 2). 

Each condition is evolved in the same circular gradient environment for 10 different runs. We incorporated the results in the previous tables Appendix \ref{tab:scal}) and Appendix \ref{tab:rob}). In the end, Hebbian learning outperforms all methods for all conditions ($p<0.01$), except the Hebbian-1.

\textcolor{black}{A particular property of Hebbian learning is the ability to propagate previous states through the dynamics of the weights. This enables a form of memory, that (we suspect) enabled the single agent to solve the gradient following task during the 1-Hebbian condition.} More interestingly, the introduction of additional members resulted in the inability to generalise to different environments, both when scaling and in the case of different environments that require flexibility. We suspect that the 1-Hebbian overfits on the memory leakage, which results in bad generalisation. Here, the absence of other members caused overfitting to the problem, as the complex nature of swarms presents itself as additional noise during evolution resulting in more robust control.

The recurrent NN reveals that state-propagation by itself is not sufficient to solve the gradient task. We suspect that the chaotic nature of both swarm behaviour and recurrent NN is detrimental to effectively evolve good weights in this condition. In contrast Hebbian learning propagates information much more gradually through its first-order dynamics which enabled the Hebbian-1 approach to succeed.

\subsection{Neural architecture design}\label{APP_NAD}
\begin{figure}[ht]
\vspace{-0.5em}
  \centering
\begin{minipage}[c]{0.995\linewidth}
\centering
% \hspace{2em}
        \begin{minipage}[t]{.3\linewidth}
            \centering
            \subfloat[1-Layer]{\includegraphics[trim={1em 0em 1em 2.2em},clip,width=\textwidth]{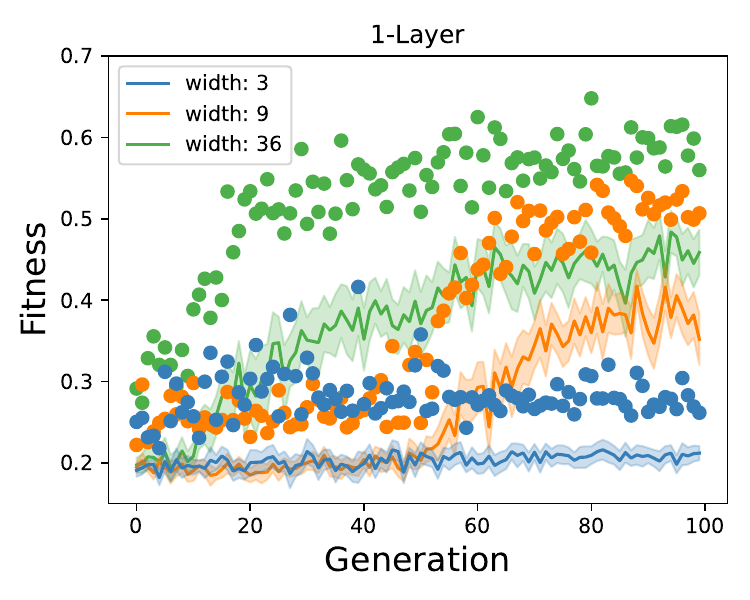}}
        \end{minipage}
        \begin{minipage}[t]{.3\linewidth}%## ROW 10x10
            \centering
            \subfloat[2-Layer]{\includegraphics[trim={1em 0em 1em 2.2em},clip, width=\textwidth]{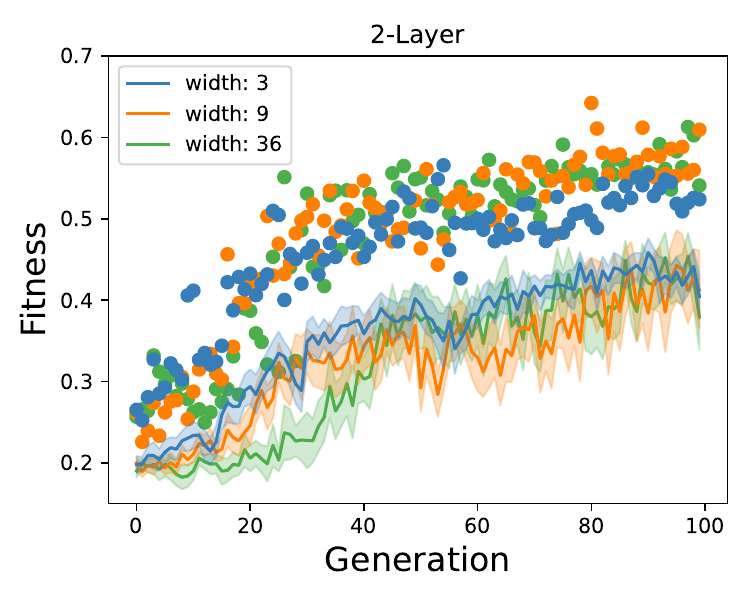}}
        \end{minipage}
% \hspace{2em}
        \begin{minipage}[t]{.3\linewidth}
            \centering
            \subfloat[3-Layer]{\includegraphics[trim={1em 0em 1em 2.2em},clip, ,width=\textwidth]{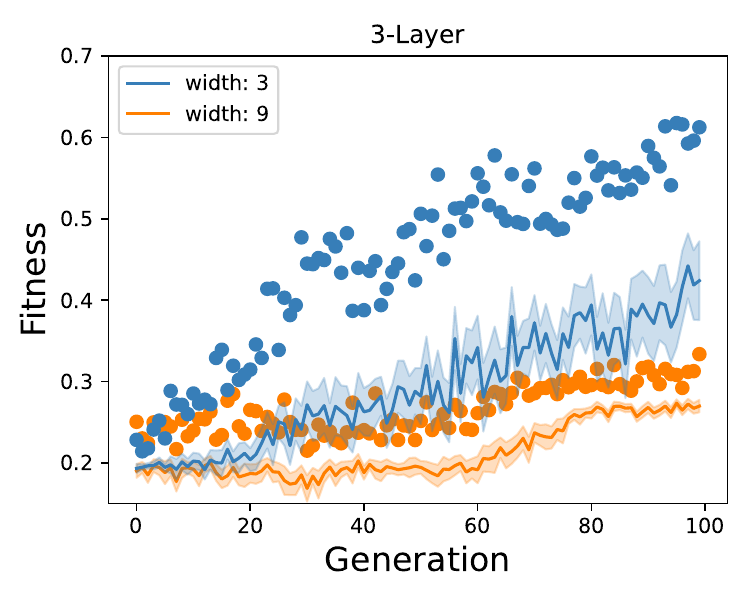}}
    \end{minipage}
\end{minipage}
% \vspace{-1em}
\caption{\small Neural architecture grid search}
\label{fig:APP_NAD}
\end{figure}

A grid-analysis was conducted for the source localisation task with different NN architectures, to estimate the sensitivity with respect to the width and depth hyper-parameters (see \autoref{fig:APP_NAD}). For this, we systematically changed the neural architecture in terms of; depth $\left[1,2,3\right]$ hidden layers (with width=9); and width $\left[3,9,36\right]$. Width 36 was chosen to obtain 720 NN weights in the Layer-2 condition (the same number of parameters as Hebbian learning). \textcolor{black}{We assess performance of different Baseline architectures (as fitness), and expressiveness (in terms of parameterisation) for fair comparison with the Hebbian controller.}. For all conditions we used the same source localisation task, with CMA-ES to evolve the weights in the network (see \autoref{sec:meth}).

From the results it is clear that for the 1-hidden layer NN architecture the performance significantly drops in performance for smaller widths (3, 9 particular). This indicates that the task might require a minimal level of nonlinearity/parametrization that is hard to capture with a single hidden layer. The 1-layer 36 width condition seems to perform similarly as all the 2-hidden layer conditions. In 2-hidden layer case we see an increase in performance where the width of the network does not seem to matter too much. Overall, the mean performance progresses similarly for all widths in the 2-layer networks, except the 36 width network takes more time to improve initially. This indicates over-parameterisation for this architecture which can slow down the evolutionary process. This trend continues for the 3-hidden layer networks where evolution requires significantly more time to optimise (we were unable to complete the 36-width layer 3 condition in a manageable time-frame). Nevertheless, the width 3 and 9 condition in the 3-layer indicate the same tendency to significantly slower convergence when increasing the parameters beyond 180 weights. 

\textcolor{black}{In the end we decide to maximize expressiveness with the 2-layer 9 width network which is among the best performing architectures among those we tested: Architectural increases in the number of weights did not meaningfully improve performance, while wall-clock time significantly increased}.

\subsection{Weight distribution}\label{APP_hists}
We took a snapshot of the distribution of the weights at constant time intervals of 150 seconds during the dynamic vs. static light experiments (see \autoref{sec:dyn_vs_stat}). As a reminder, the best Hebbian controller from \autoref{sec:res} is retested in the same circular gradient environment where we perturb the position of the light source after 300 seconds of deployment (600 seconds are simulated in total). The static condition does not involve this perturbation. We shows the distribution of weights at the $\left[0, 150, 300, 450, 600\right]$ time intervals below:

\begin{figure}[ht]
\begin{minipage}[c]{0.995\linewidth}
\centering
        \begin{minipage}[t]{.3\linewidth}
            \centering
            \subfloat[t0]{\includegraphics[trim={1em 0em 1em 2.2em},clip,width=\textwidth]{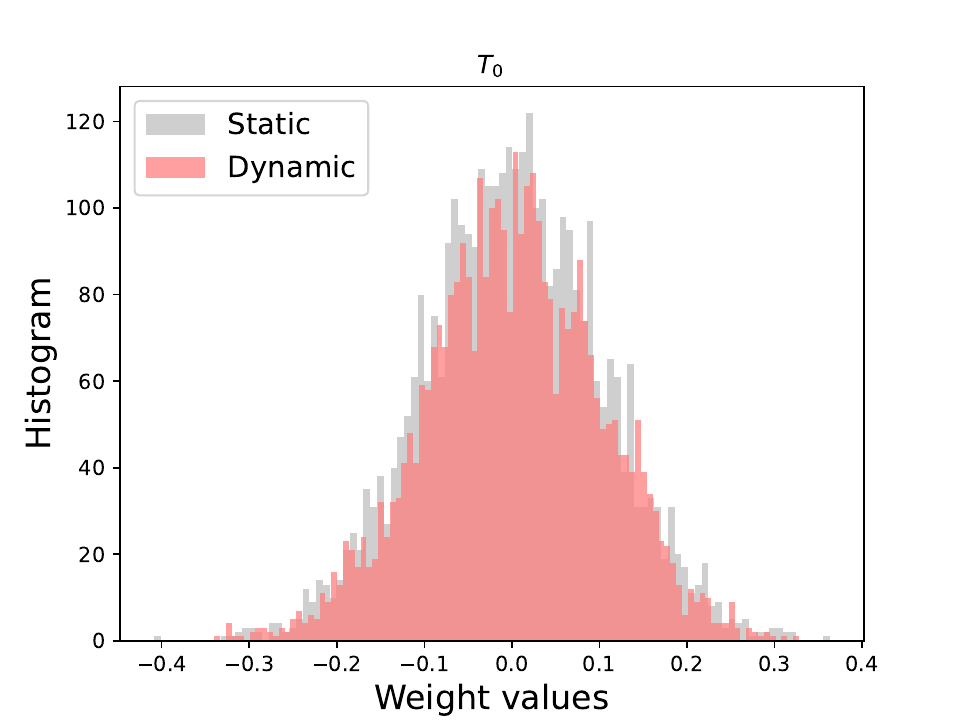}}
        \end{minipage}
        \begin{minipage}[t]{.3\linewidth}%## ROW 10x10
            \centering
            \subfloat[t150]{\includegraphics[trim={1em 0em 1em 2.2em},clip, width=\textwidth]{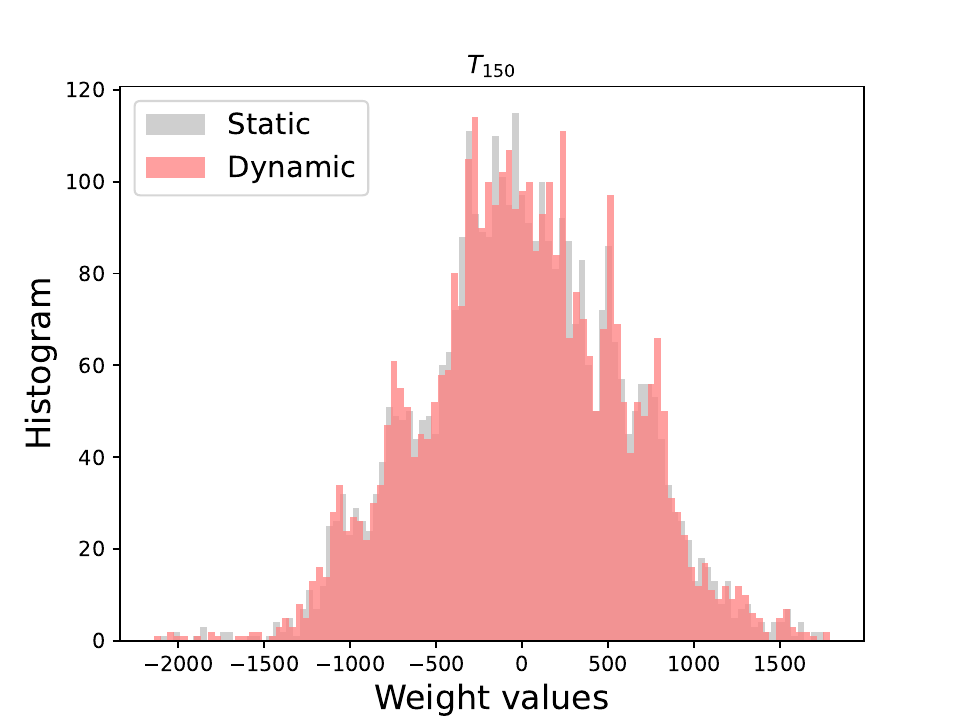}}
        \end{minipage}
        \begin{minipage}[t]{.3\linewidth}
            \centering
            \subfloat[t300]{\includegraphics[trim={1em 0em 1em 2.2em},clip, ,width=\textwidth]{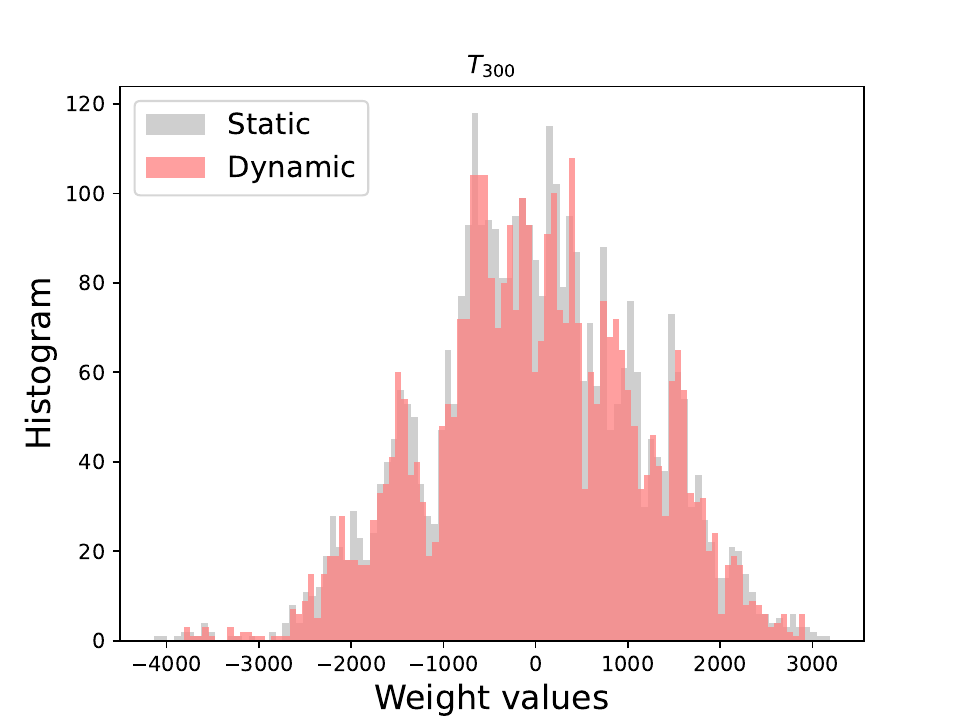}}
    \end{minipage}\\
% \hspace{2em}
        \begin{minipage}[t]{.4\linewidth}
            \centering
            \subfloat[t450]{\includegraphics[trim={1em 0em 1em 2.2em},clip, ,width=\textwidth]{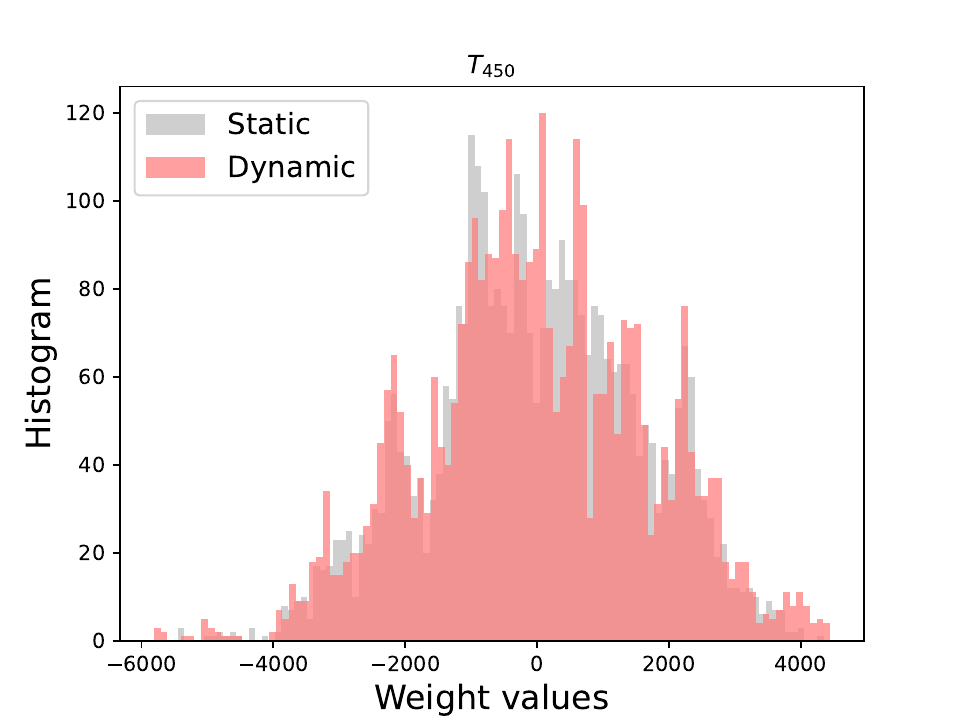}}
    \end{minipage}
        \begin{minipage}[t]{.4\linewidth}
            \centering
            \subfloat[t600]{\includegraphics[trim={1em 0em 1em 2.2em},clip, ,width=\textwidth]{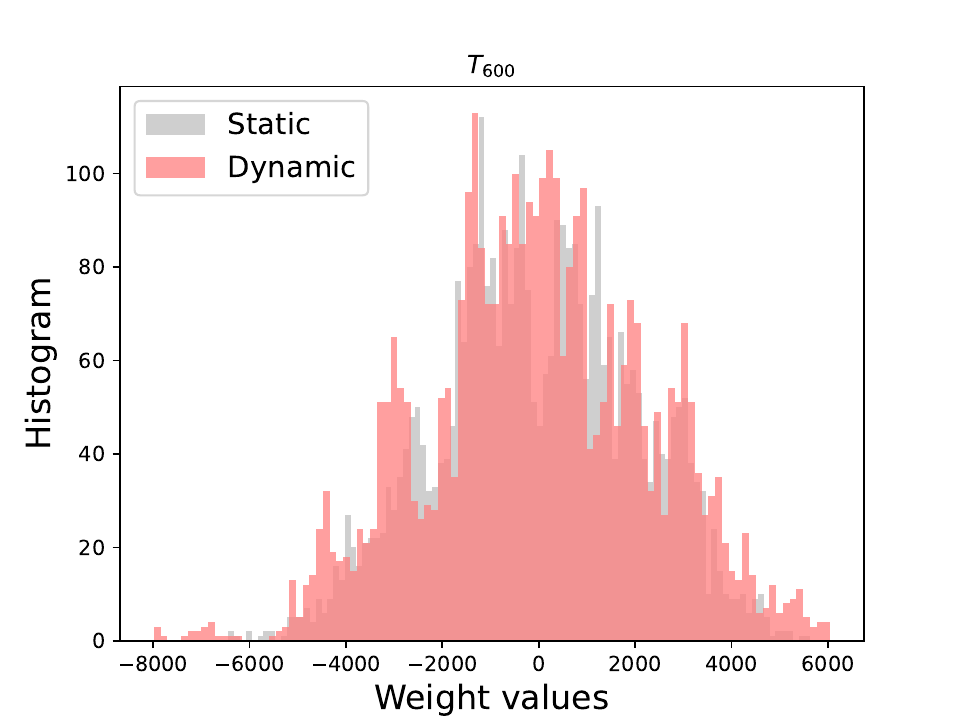}}
    \end{minipage}
\end{minipage}
\caption{\small Weight distribution during deployment}
\label{fig:APP_hist}
\end{figure}

From 0 to 300 seconds the weights are similarly distributed between the static and dynamic environment. This is not surprising as up until this point both dynamic and static conditions are identical. At 450 seconds a difference start to emerge, with more weights distributing around zero in the dynamic case. This is even more visible at 600 seconds where the static condition has a clearly lower density around zero with respect to the dynamic condition. These findings are consistent with the reduced weight variability found after perturbation (see \autoref{fig:dyn_v_stat}e) for the dynamic case. 

%%=============================================%%
%% For submissions to Nature Portfolio Journals %%
%% please use the heading ``Extended Data''.   %%
%%=============================================%%

%%=============================================================%%
%% Sample for another appendix section			       %%
%%=============================================================%%

%% \section{Example of another appendix section}\label{secA2}%
%% Appendices may be used for helpful, supporting or essential material that would otherwise 
%% clutter, break up or be distracting to the text. Appendices can consist of sections, figures, 
%% tables and equations etc.

\end{appendices}

\end{document}